\documentclass[letterpaper, 10 pt, conference]{ieeeconf}  

\IEEEoverridecommandlockouts                              

\usepackage{cite}
\usepackage{graphicx,epstopdf,epsfig}
\graphicspath{{figures/}}
\DeclareGraphicsExtensions{.pdf}
\usepackage{amsmath}
\usepackage{amssymb}

\usepackage{array}
\usepackage{fixltx2e}
\usepackage{url}
\usepackage{booktabs}
\usepackage{float}
\usepackage{hyperref}
\usepackage[caption=false,font=scriptsize,labelfont=sf,textfont=sf]{subfig}
\title{\LARGE \bf
	Profiling Visual Dynamic Complexity Using a Bio-Robotic Approach
}

\author{
	Qinbing Fu$^{1,2}$, Tian Liu$^{2}$, Xuelong Sun$^{2}$, Huatian Wang$^{2}$, Jigen Peng$^{1,3}$, Shigang Yue$^{1,2}$, Cheng Hu$^{1,2}$
\thanks{
	$^{1}$Machine Life and Intelligence Research Centre, Guangzhou University, Guangzhou, 510006, China
}
\thanks{
	$^{2}$School of Computer Science, University of Lincoln, Lincoln, LN6 7TS, United Kingdom
}
\thanks{
	$^{3}$School of Mathematics and Information Science, Guangzhou University, Guangzhou, 510006, China
}
\thanks{
	Q. Fu, T. Liu, and X. Sun are joint first authors. 
	Correspondence: \url{{chu,syue}@lincoln.ac.uk} \url{qinbingfu87@gmail.com}
}
}

\begin{document}

\maketitle
\thispagestyle{empty}
\pagestyle{empty}

\begin{abstract}
	
Visual dynamic complexity is a ubiquitous, hidden attribute of the visual world that every dynamic vision system is faced with. 
However, it is implicit and intractable which has never been quantitatively described due to the difficulty in defending temporal features correlated to spatial image complexity. 
To fill this vacancy, we propose a novel bio-robotic approach to profile visual dynamic complexity which can be used as a new metric. 
Here we apply a state-of-the-art brain-inspired motion detection neural network model to explicitly profile such complexity associated with spatial-temporal frequency (SF-TF) of visual scene. 
This model is for the first time implemented in an autonomous micro-mobile robot which navigates freely in an arena with visual walls displaying moving sine-wave grating or cluttered natural scene. 
The neural dynamic response can make reasonable prediction on surrounding complexity since it can be mapped monotonically to varying SF-TF of visual scene. 
The experiments show this approach is flexible to different visual scenes for profiling the dynamic complexity. 
We also use this metric as a predictor to investigate the boundary of another collision detection visual system performing in changing environment with increasing dynamic complexity. 
This research demonstrates a new paradigm of using biologically plausible visual processing scheme to estimate dynamic complexity of visual scene from both spatial and temporal perspectives, which could be beneficial to predicting input complexity when evaluating dynamic vision systems.

\end{abstract}
\section{Introduction}
\label{Sec: introduction}

When assessing dynamic vision systems, it would be more convincing to carry out the research progressively by using a range of input visual stimuli signals from simple, normally synthesised with clean backgrounds to handcrafted samples with more complex backgrounds like natural cluttered visual scene, finally to unstructured indoor and outdoor real world scenarios. 
The moving object and background are all controllable in handcrafted testing visual scene; in this case, the visual system models can be systematically investigated as desired. 
The real world visual scenes are nevertheless intractable; the models usually perform more randomly.

Many visual system modelling studies insisted on working effectively and robustly in complex and dynamic environments. 
However, none of those have quantitatively estimated the complexity of dynamic visual scene leaving the said ``dynamic-complex" implicit and fully empirical. 
Accordingly, it could be hard to define or find the boundary of visual system models challenged by unstructured, real physical scene. 
Under constrained parameters, the models could also be overestimated without quantitative analysis on dynamic complexity of visual scene.

In this regard, we reflected on how to estimate the visual dynamic complexity with quantitative description. 
Image processing researchers have presented some prominent methods on evaluating the spatial image complexity. 
Yu and Winkler reviewed relevant methods including compression-based complexity, spatial information indicated by edge energy \cite{Yu-2013-image-complexity}. 
Some methods have been requesting observers to rate the perceived complexity of images \cite{Wu-2006-complexity,Chikhman-2012-complexity,Murguia-2005-complexity}. 
There have been complexity-based similarity measures used to understand content-based image recognition problems \cite{Perkio-2009-complexity}, cluster and classify images \cite{Guha-2012-complexity}, aesthetic pictures \cite{Romero-2012-complexity}. 
Briefly speaking, the definition of image complexity is still in debate, as most methods are based on subjective measures that are inconsistent. 
On the aspect of objective measures, Corchs et al. proposed a new complexity measurement by linearly combining different image features including spatial, frequency and colour properties \cite{Corchs-2016-Complexity-Perception}. 
More recently, there were also methods focusing on quantifying the complexity of black-and-white images based on the detection of typical length scales \cite{Zanette-2018}, as well as modelling visual complexity via image features fusion of multiple kernels \cite{Lozano-2019-visual-complexity}.

However, to the best of our knowledge, none of these methods have analysed visual complexity from a temporal perspective. 
A potential challenge is how to defend correlated features in both space and time scales. 
To this end, biological visual neural systems are capable of measuring then adapting to changing environment very quickly and robustly. 
Moreover, the neural circuits balance well performance and power consumption where some neural activities could be flexible to rise up in relatively more complex dynamic environment whilst resting in easy conditions. 
The underlying mechanisms on measuring visual dynamic complexity for such self-adaptation have not yet been understood.

Neuroscientists have found the spatial-temporal frequency (SF-TF) of movements are important attributes of dynamic visual scene \cite{Nature-1997-spatial-temporal-frequency}. 
Visual neurons show selectivity to different frequencies corresponding to distinct responsive activities \cite{Foster-1984-ST-frequency-selectivity,Priebe-2006-st-frequency-speed-tuning}. 
Besides that, a vast majority of physiological works apply sine-wave gratings as input stimuli to neural circuits and pathways \cite{Borst2011(review-motion)}, as the gratings can precisely reflect the attributes of visual scene in both SF-TF domains. 
Accordingly, visual sensory systems are likely encoding such neural dynamic characteristics for complexity measurement.

Recently, a brain-inspired motion detection neural network model was proposed which can decode spatiotemporal attributes of visual scene into interpretable angular velocity, largely independent of spatial frequency and contrast information \cite{AVDM-NN}. 
This fits well with our objective to further develop a metric to involve also temporal features coupled with spatial image complexity. 
Firstly, we investigate the relation between model response and visual dynamic complexity associated with both SF-TF of visual scene. 
Importantly, we have found a mathematically monotonic relation indicating that higher complexity (SF or TF) of visual scene brings about stronger model dynamic response. 
This can be established as a new metric to estimate visual complexity from both spatial and temporal perspectives. 
After that, we implement this approach into embedded vision of an autonomous micro-mobile robot. 
During visual navigation in an arena encompassed by visually dynamic walls displaying either grating or drifting natural scene at varying frequencies, such complexity can be profiled by on-line visual processing, effectively and efficiently, in different situations. 
Finally, we demonstrate a practical use of this new metric as a predictor to find the boundary of a collision detection visual system in changing environment with increasing complexity. 
The bio-robotic experiments demonstrate this biologically plausible metric is feasible and computing efficient to profile visual dynamic complexity which could be beneficial to evaluating other dynamic vision systems as a predictor of input complexity. 
This approach could also be scalable to a variety of robotic visual tasks to measure changing environment.

The rest of this paper is organised as follows. 
Section \ref{Sec: methods} presents the methods and materials. 
Section \ref{Sec: evaluation} shows the metric validation through off-line simulation and on-line robot tests. 
Section \ref{Sec: profile} exhibits the profiled complexity through robot visual navigation. 
Section \ref{Sec: investigation} demonstrates a valid use of this new metric. 
Section \ref{Sec: conclusion} concludes this paper.

\section{Methods and Materials}
\label{Sec: methods}


\subsection{Model}

\begin{figure}[t]
	\centering
	\includegraphics[width=0.49\textwidth]{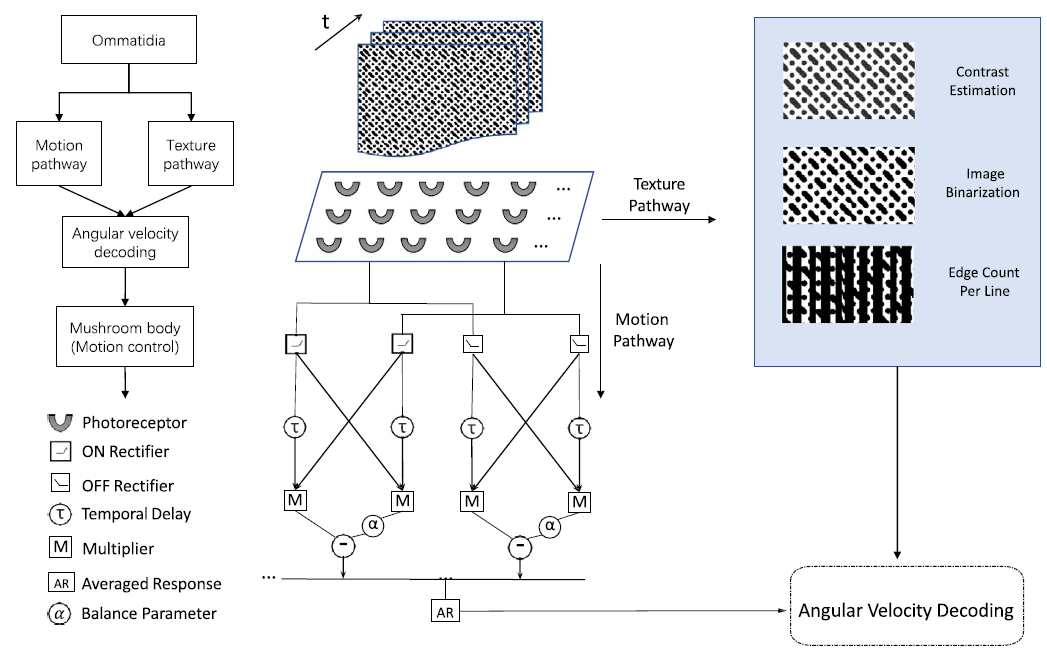}
	\caption{
		Illustration of the AVDM model adapted from \cite{AVDM-NN} for profiling visual dynamic complexity in this research. 
		The neural network model mainly consists of two pathways, the motion and the texture pathways to encode spatiotemporal features. 
		At the output layer, the coupled signals are interpreted to indicate visual angular velocity of movements.
	}
	\label{Fig: AVDM}
\end{figure}

This proposed metric is established by a biologically plausible visual processing scheme implemented in a vision-based micro-mobile robot. 
First of all, the dynamic vision system is a recently published brain-inspired motion detection neural network model called ``angular velocity decoding model" (AVDM) mimicking honeybee's motion vision neural pathways to transform spatiotemporal image features into interpretable visual angular velocity \cite{AVDM-NN}, as shown in Fig. \ref{Fig: AVDM}. 
Previously, the AVDM was only tested by handcrafted visual stimuli. 
In this research, it is for the first time implemented in robot vision. 
The emphasis herein is laid on further exploring its efficacy in robot visual navigation, and potential on profiling dynamic complexity. 
There have been many similar visual system models that work effectively to encode diverse motion cues \cite{Fu-ALife-review}. 
The motivation to select the AVDM to estimate visual dynamic complexity is its large independence on spatial frequency and image contrast compared to the current repository.

To be more specific, the AVDM mainly consists of two separate visual neural pathways. 
The texture pathway measures image complexity in space only, which works effectively to deal with SF and image contrast. 
The motion pathway focuses on TF to correlate features in time. 
Here we do not repeat the algorithms in the two visual pathways, but emphasise the last output ``decoding layer" which is crucial to estimate the dynamic complexity associated with SF-TF of visual scene. 
Given the outputs from texture ($\lambda$) and motion ($R$) pathways, the AVDM's initial idea is to decode the angular velocity ($\omega$) from the response $R(\omega,\lambda)$ using an approximation method. 
Consequently, the actual angular velocity can be approximated combining a power function of texture pathway output, and a square root function of motion pathway response, as follows: 
\begin{equation}
\omega \approx \hat{a} \lambda^{\hat{b}} \frac{1+\hat{C}}{2\hat{C}} \sqrt{R},
\label{Eq: av-approximation}
\end{equation}
where $\hat{C}$ is the estimated spatial contrast. 
Parameters $\hat{a}$ and $\hat{b}$ can be learnt by minimizing the difference from ground truth, iteratively as: 
\begin{equation}
(\hat{a},\hat{b})=\text{arg}\underset{a,b}{\text{min}}(\omega - a \lambda^b \frac{1+C}{2C} \sqrt{R(\omega,\lambda)}).
\label{Eq: av-optimisation}
\end{equation}

\begin{figure}[t]
	\centering
	\includegraphics[width=0.49\textwidth]{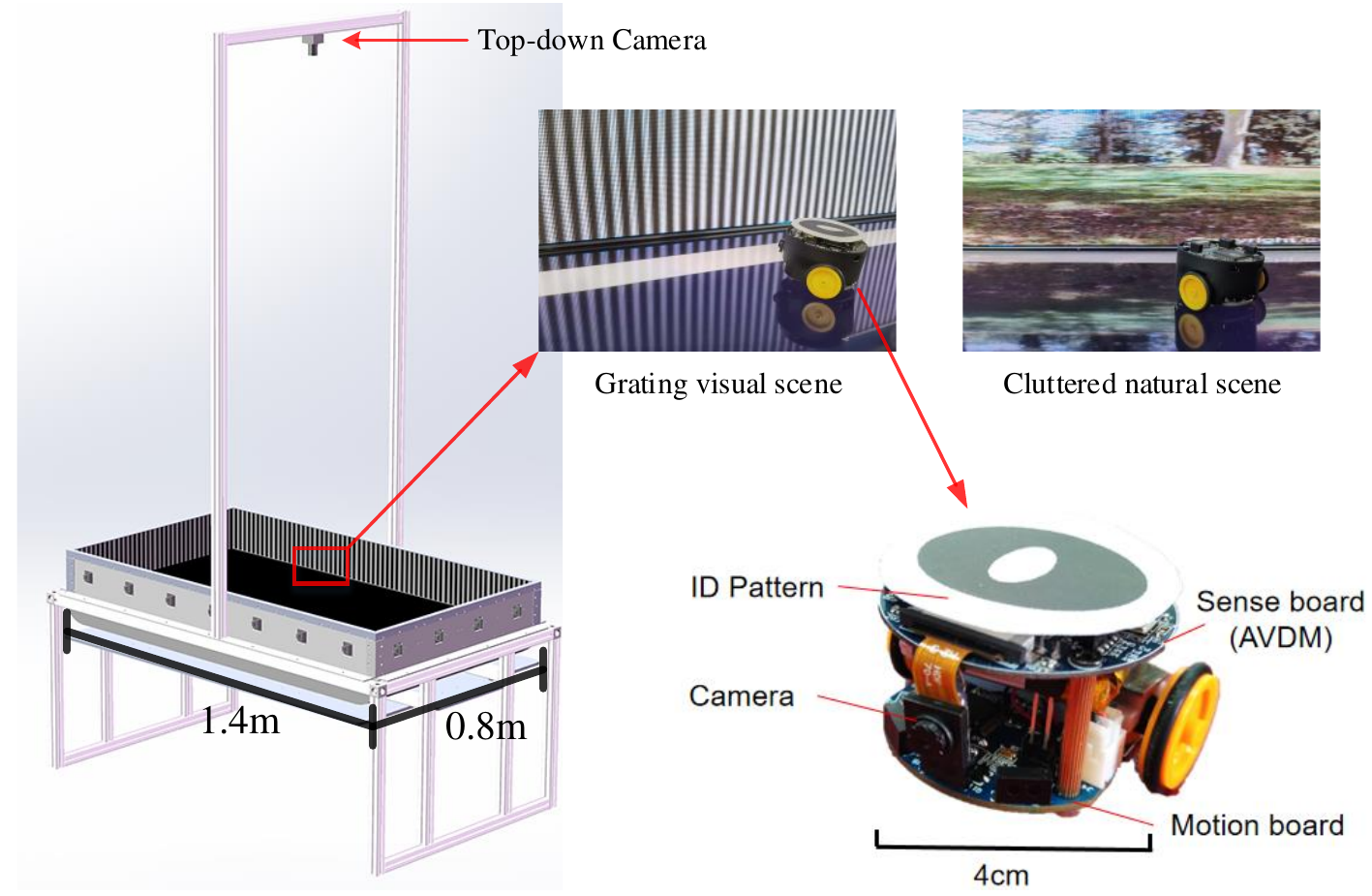}
	\caption{
		Illustration of the robot arena with visual walls displaying different dynamic visual scenes including grating and cluttered natural scenes. 
		The micro-robot is used for profiling the visual dynamic complexity.
	}
	\label{Fig: arena-robot}
	\vspace{-10pt}
\end{figure}

\begin{figure}[t]
	\centering
	\subfloat[moving grating stimuli in time]{\includegraphics[width=0.25\textwidth]{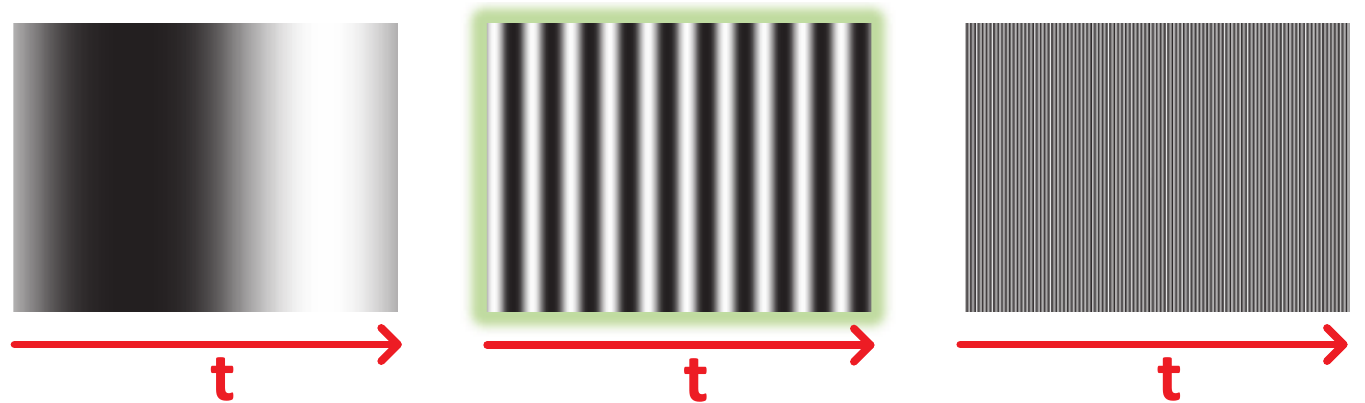}
		\label{Fig: grating-stimuli}}
	\vfil
	\vspace{-5pt}
	\subfloat[SF=1upc, dynamic TF]{\includegraphics[width=0.23\textwidth]{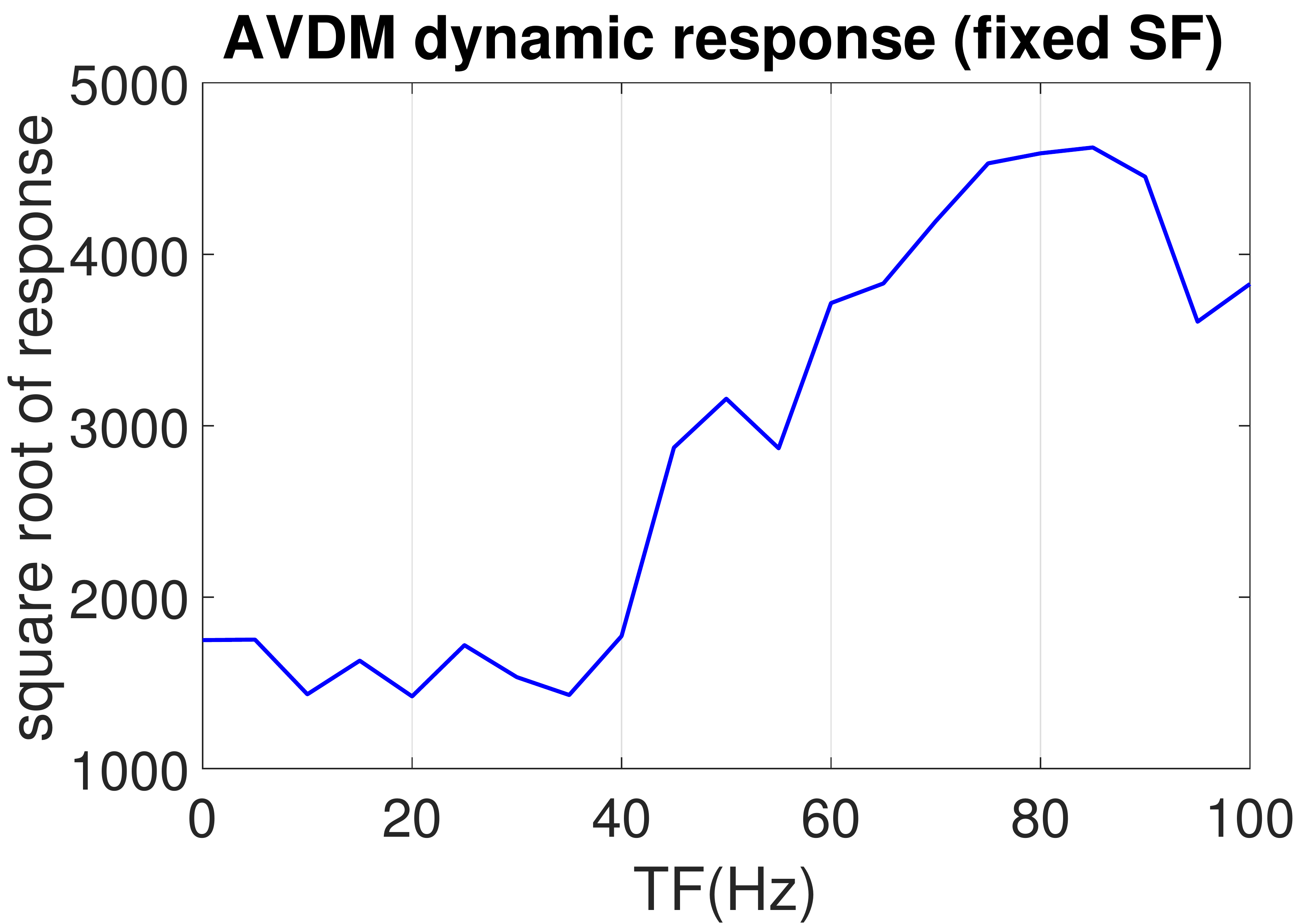}
		\label{Fig: avdm-sf1}}
	\hfil
	\subfloat[TF=1Hz, dynamic SF]{\includegraphics[width=0.23\textwidth]{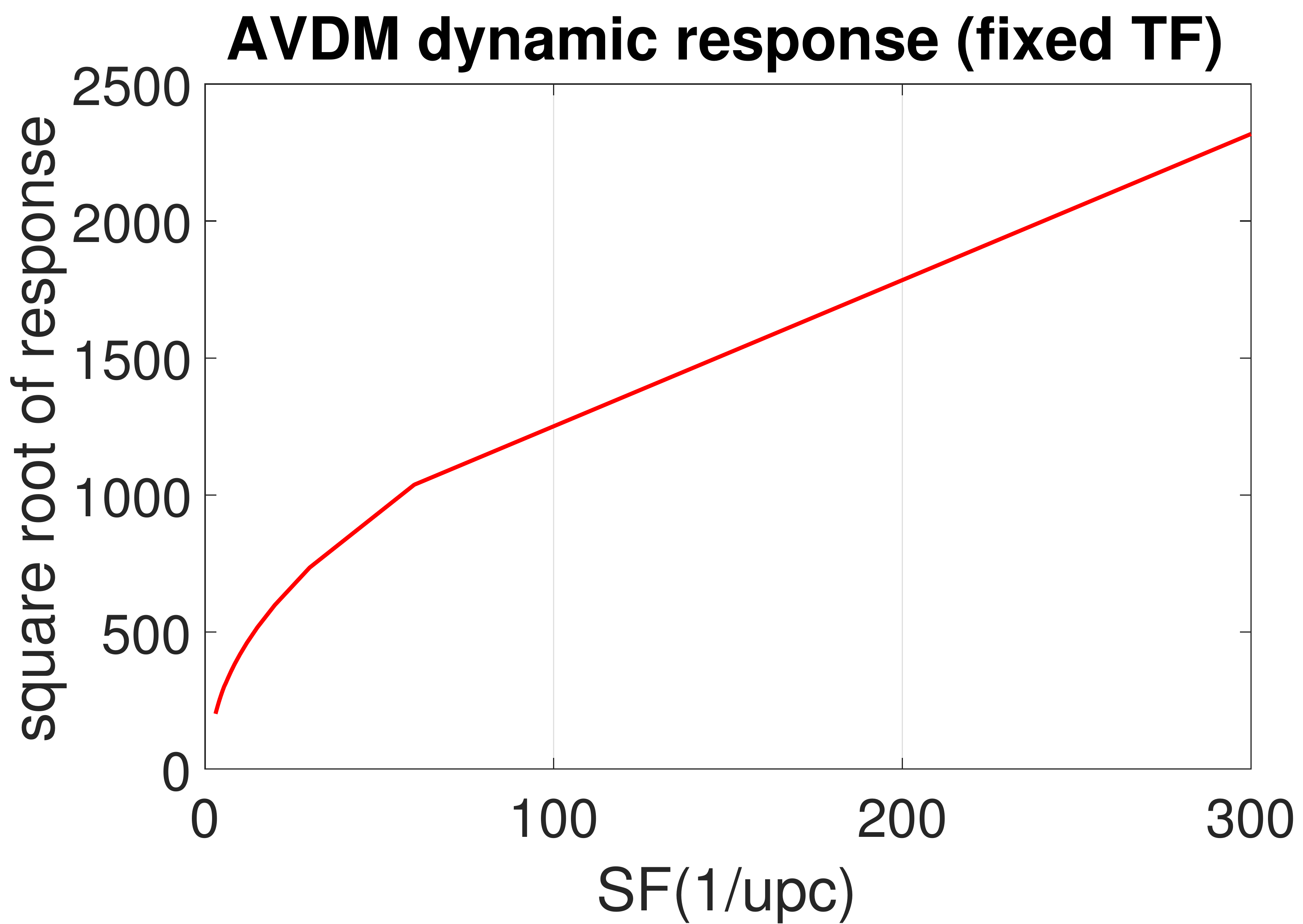}
		\label{Fig: avdm-tf1}}
	\vfil
	\vspace{-5pt}
	\subfloat[SF=10upc, dynamic TF]{\includegraphics[width=0.23\textwidth]{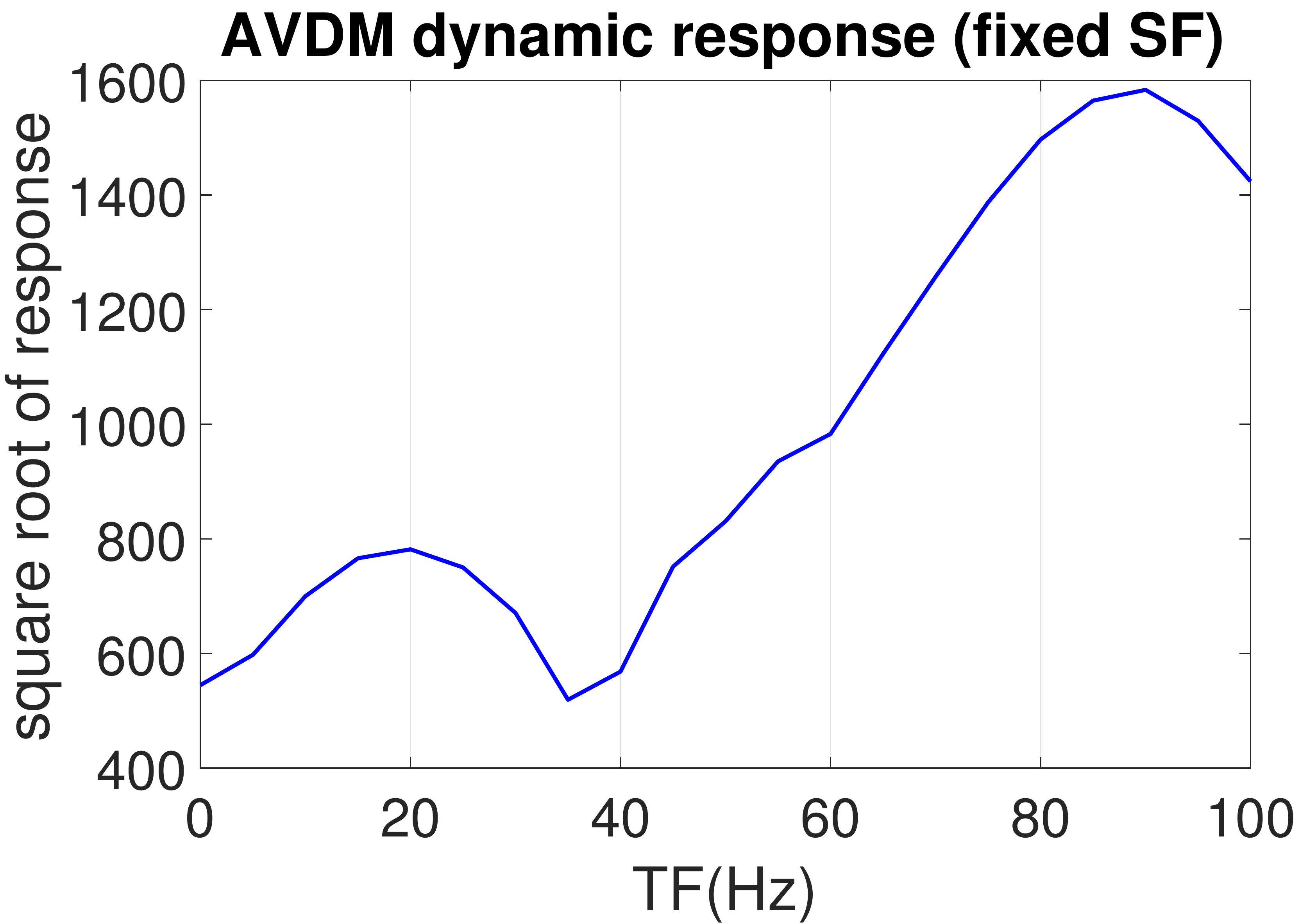}
		\label{Fig: avdm-sf10}}
	\hfil
	\subfloat[TF=5Hz, dynamic SF]{\includegraphics[width=0.23\textwidth]{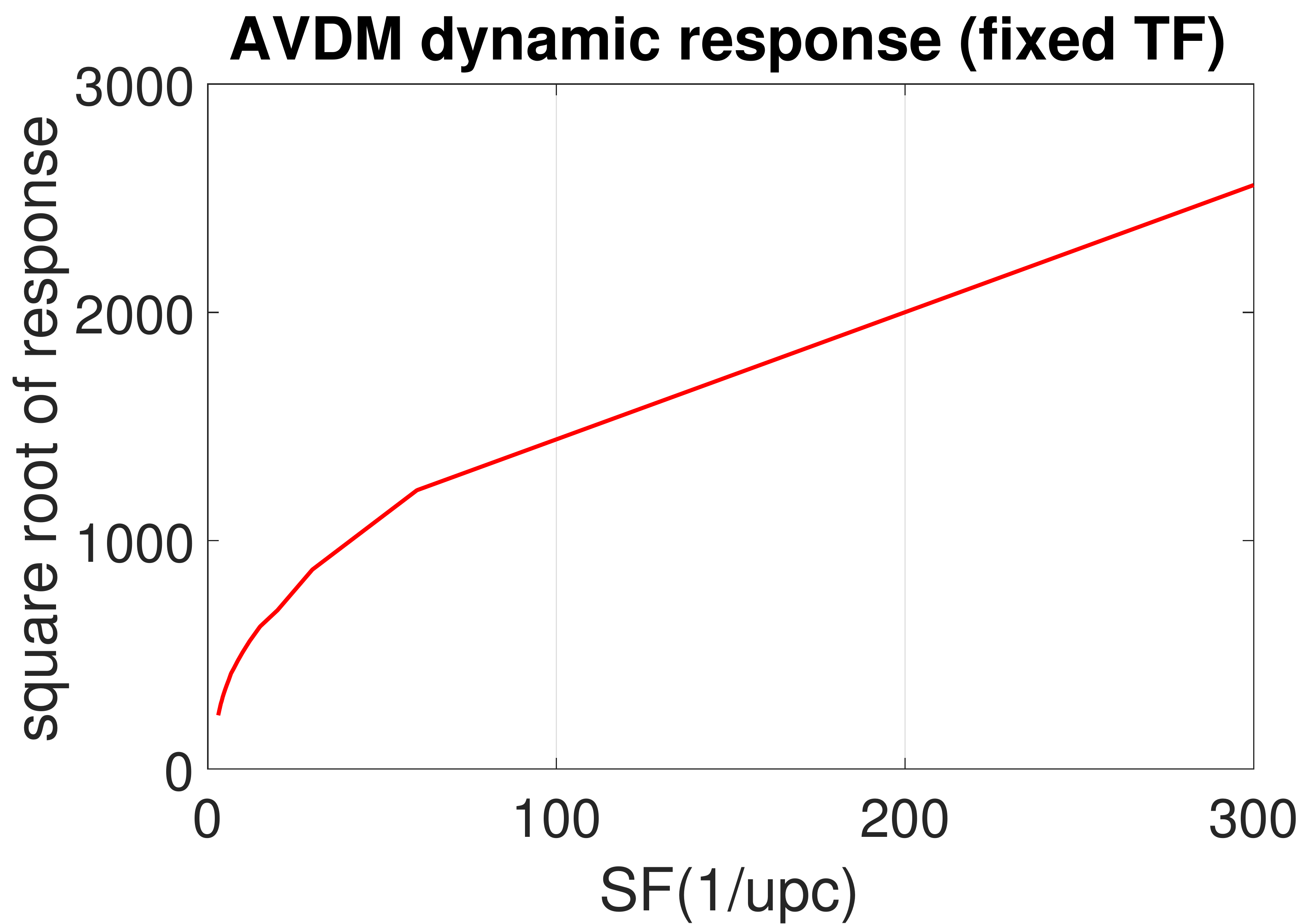}
		\label{Fig: avdm-tf5}}
	\vfil
	\vspace{-5pt}
	\subfloat[SF=50upc, dynamic TF]{\includegraphics[width=0.23\textwidth]{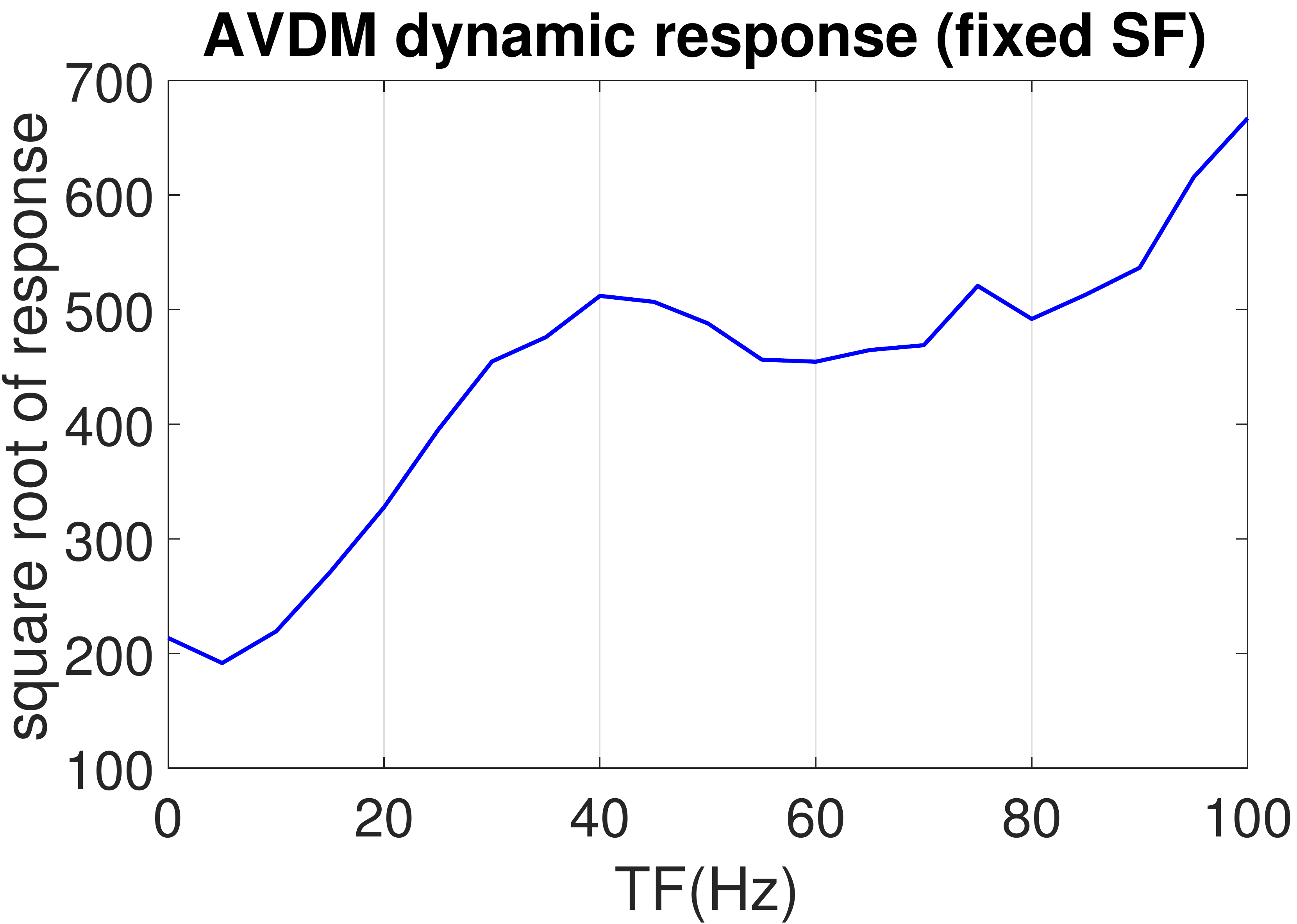}
		\label{Fig: avdm-sf50}}
	\hfil
	\subfloat[TF=50Hz, dynamic SF]{\includegraphics[width=0.23\textwidth]{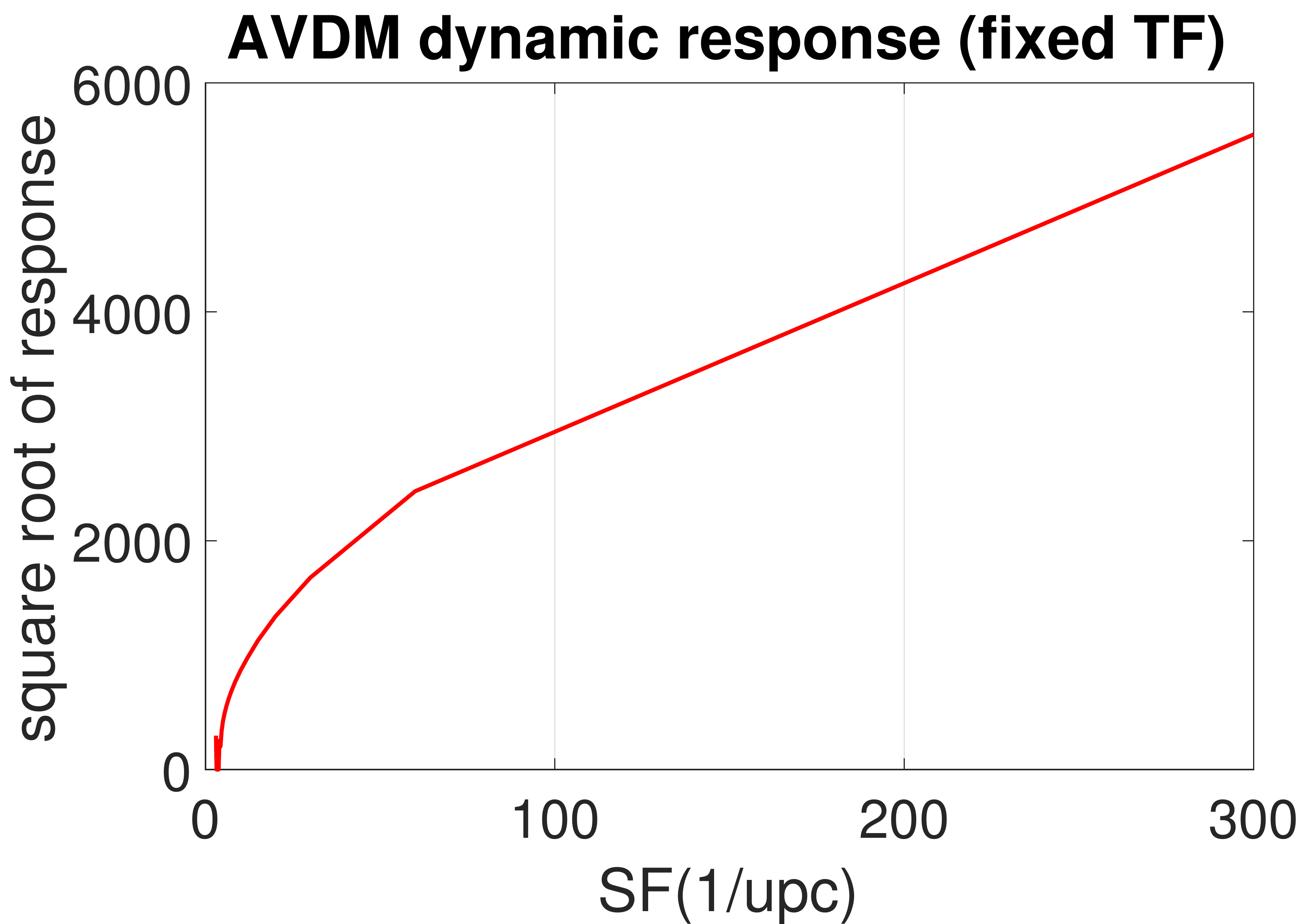}
		\label{Fig: avdm-tf50}}
	\caption{
		Dynamic response of AVDM challenged by different SF-TF gratings. 
		Here the SF metric is short for ``units per circle".
	}
	\label{Fig: simulation results}
	\vspace{-10pt}
\end{figure}

\begin{figure}[t]
	\centering
	\subfloat[three camera snapshots during approaching]{\includegraphics[width=0.3\textwidth]{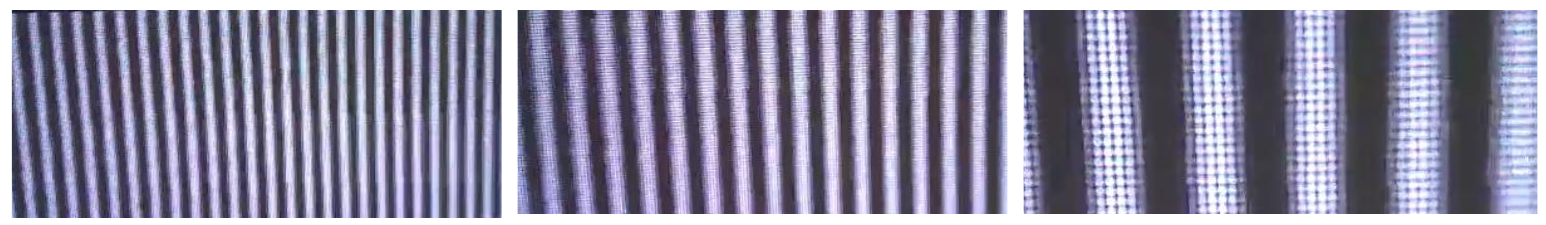}
		\label{Fig: camera snapshots}}
	\vfil
	\vspace{-5pt}
	\subfloat[grating wall TF=0Hz]{\includegraphics[width=0.18\textwidth]{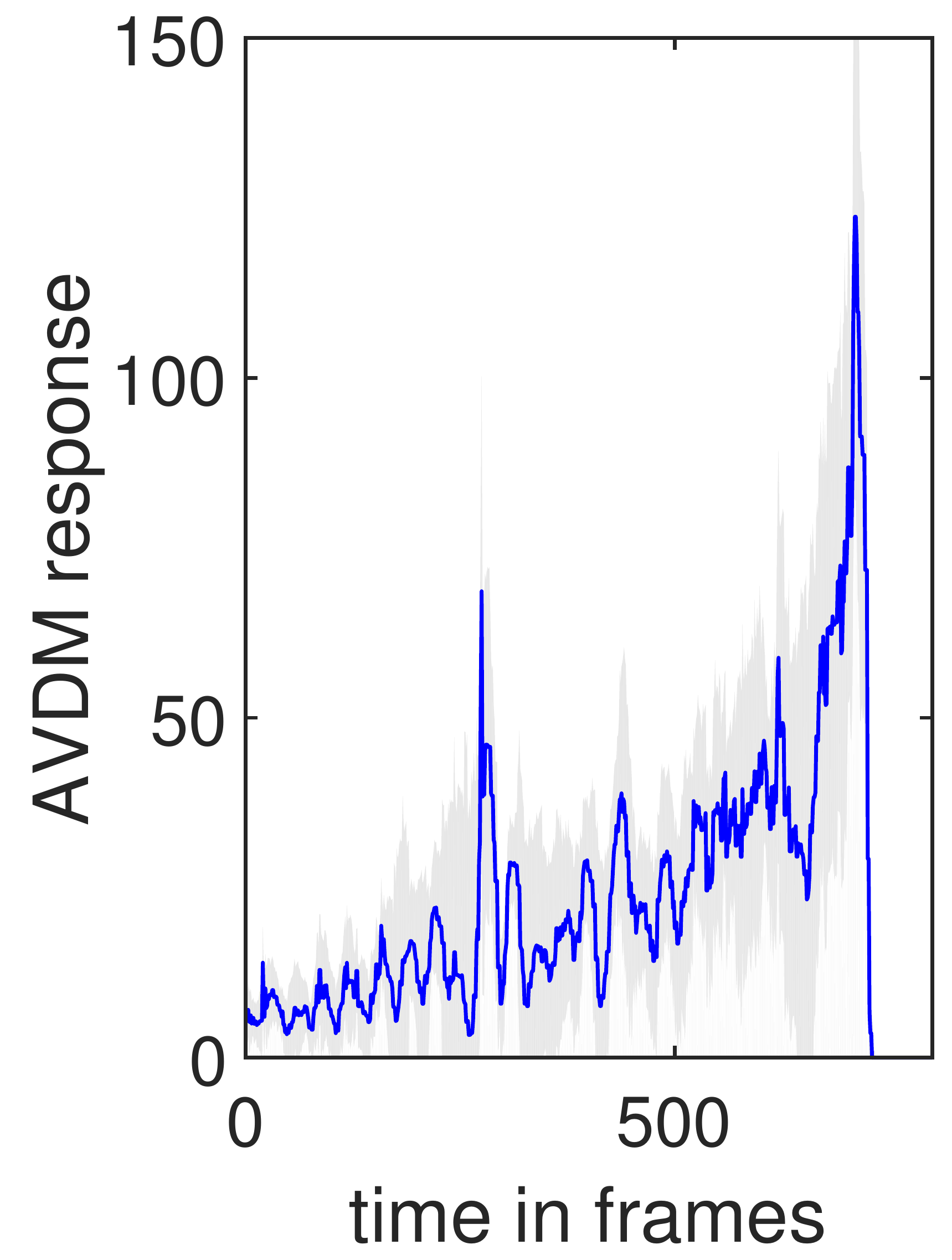}
		\label{Fig: robot-tf0}}
	\hfil
	\subfloat[grating wall TF=1Hz]{\includegraphics[width=0.18\textwidth]{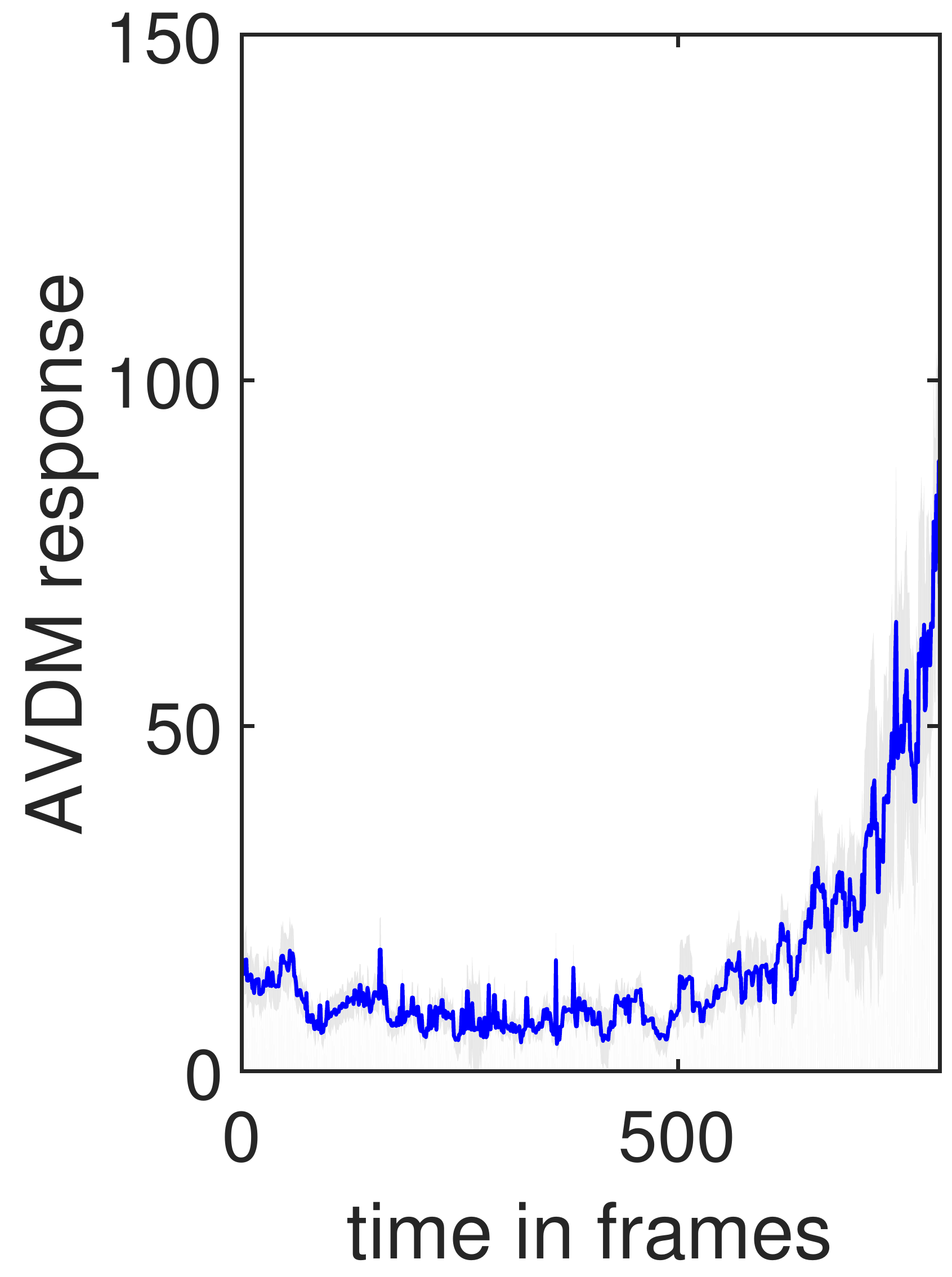}
		\label{Fig: robot-tf1}}
	\vfil
	\vspace{-5pt}
	\subfloat[grating wall TF=2Hz]{\includegraphics[width=0.18\textwidth]{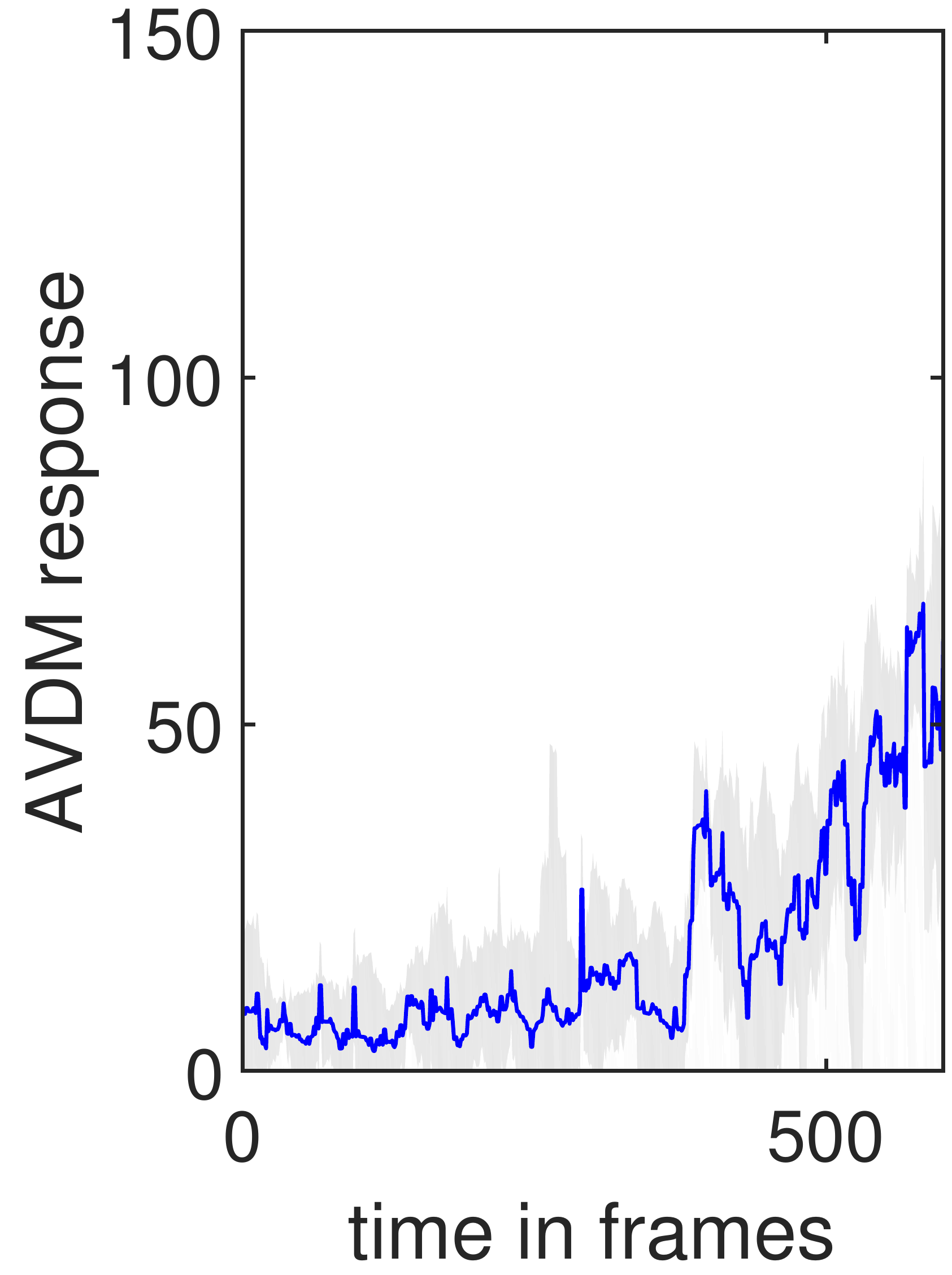}
		\label{Fig: robot-tf2}}
	\hfil
	\subfloat[grating wall TF=3Hz]{\includegraphics[width=0.18\textwidth]{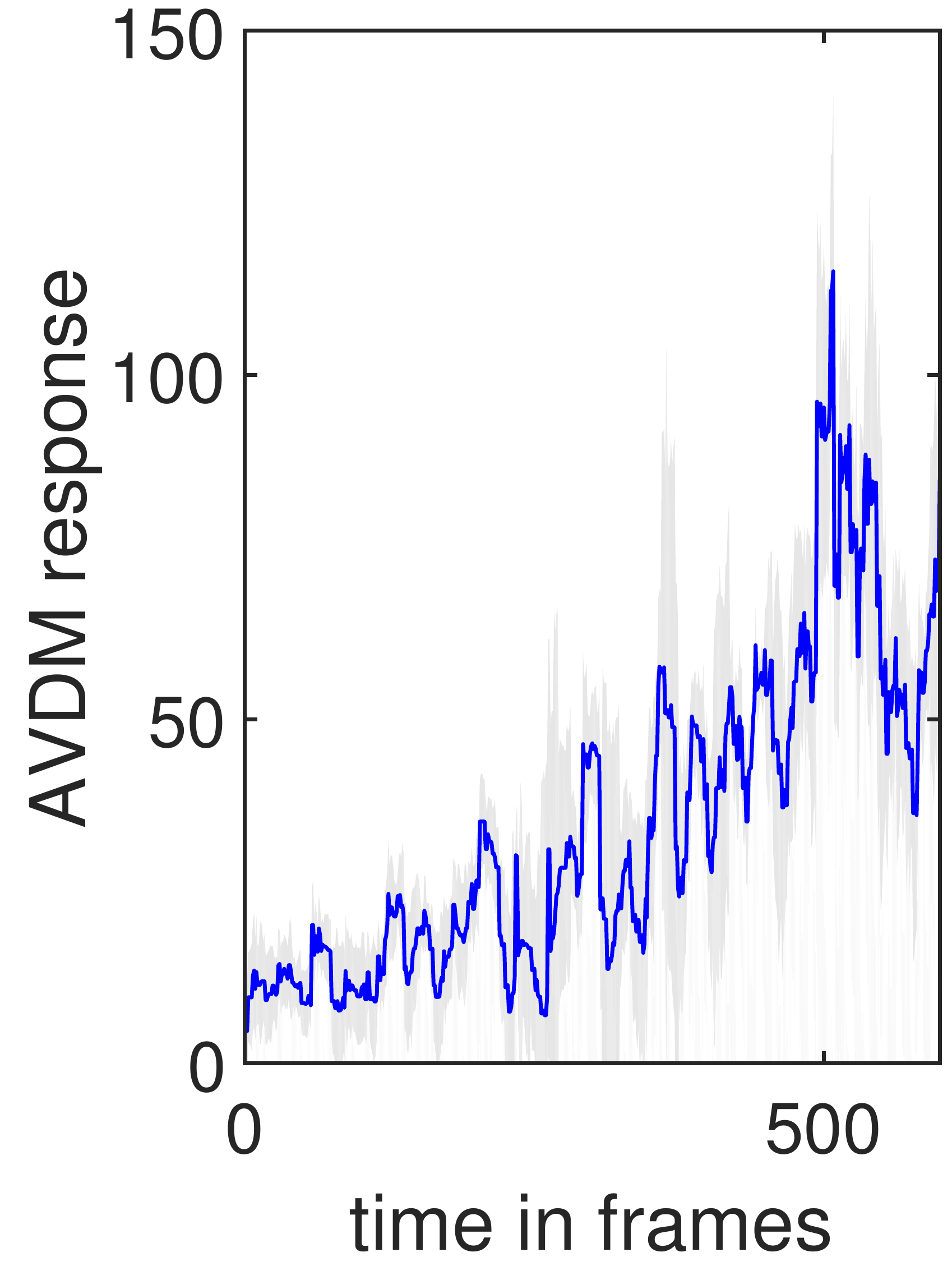}
		\label{Fig: robot-tf3}}
	\caption{
		Dynamic response of AVDM during approaching the grating wall: each process is repeated ten times (variance in grey shadow).
	}
	\label{Fig: robotic evaluation}
	\vspace{-10pt}
\end{figure}

\begin{figure}[t]
	\centering
	\subfloat[profile grating visual scene]{\includegraphics[width=0.23\textwidth]{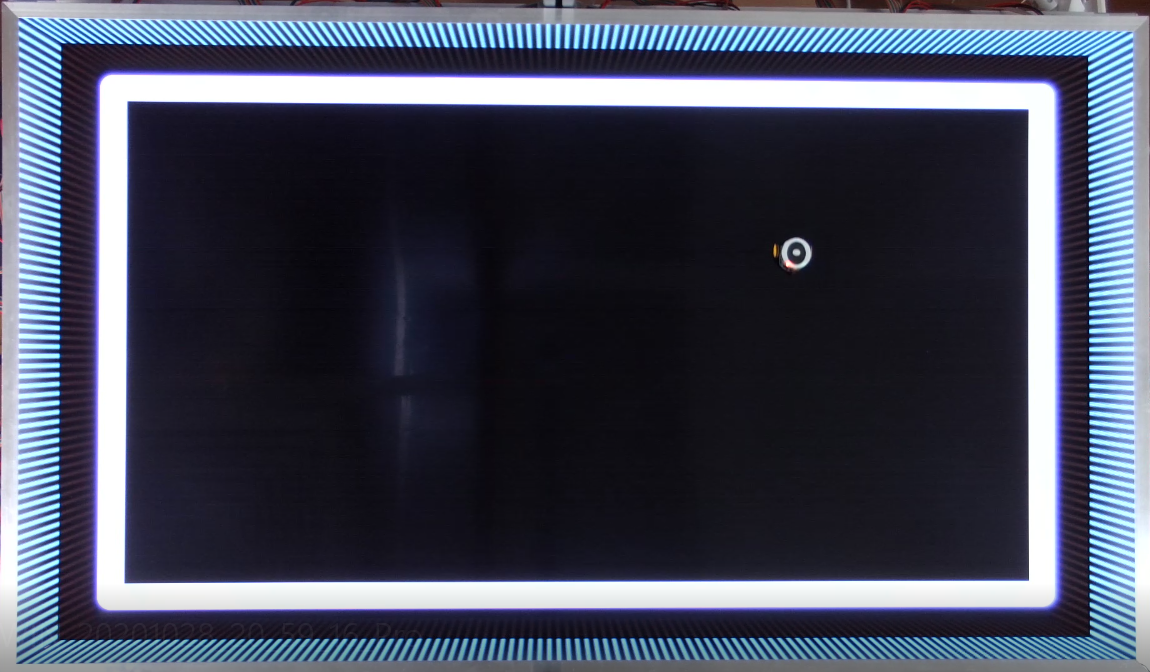}
		\label{Fig: profile setting-grating}}
	\hfil
	\subfloat[profile cluttered natural scene]{\includegraphics[width=0.23\textwidth]{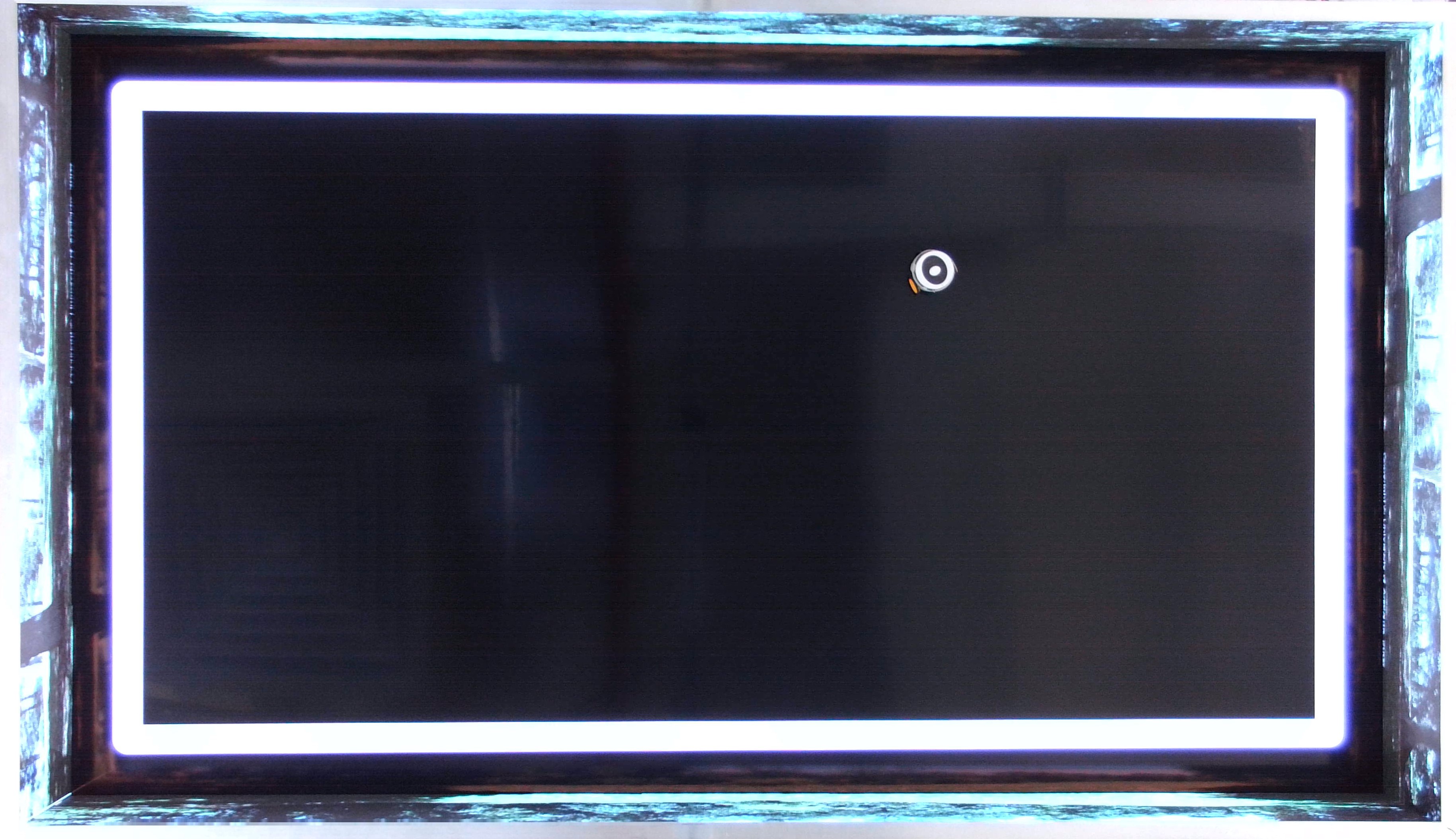}
		\label{Fig: profile setting-nature}}
	\caption{
		Arena settings (from top-down view) for profiling visual dynamic complexity in (a) grating, (b) cluttered natural scenes.
	}
	\label{Fig: arena settings}
	\vspace{-10pt}
\end{figure}

\subsection{Configuration}

Next, we introduce the configuration of model, robot, and arena. 
The parameters setting of the AVDM on embedded vision are consistent with \cite{AVDM-NN}. 
As exhibited in Fig. \ref{Fig: arena-robot}, the micro-mobile robot is called ``\textit{Colias-IV}", which is an autonomous, vision-based, multi-sensor mobile platform \cite{Hu-TAROS(Colias-IV)}. 
It has been successfully employed for research into swarm robotics and social insects \cite{ColCOSPhi-ICARM-2019}, and widely utilised in bio-inspired vision systems study \cite{Fu-TAROS(review)}. 
The detailed robot configuration can be found in \cite{Hu-TAROS(Colias-IV)}. 
Note that the robot motion and control strategy is out of the scope of this study. 
In all types of tests, we set the forward linear speed constant at approximately 0.08m/s. 
The turning angle is 80$\sim$100 degrees randomly to right or left. 
The sampling frequency of robot for image processing is maintained at around 33Hz.

The arena used in this research has been upgraded with controllable, LED-displaying walls in comparison with our previous arena experimentation with static, patterned walls \cite{Fu-2020-Access}. 
As shown in Fig. \ref{Fig: arena-robot}, with a top-down facing camera system localising the ID-specific pattern on top of robot in real time, we can obtain on-line data consisting of time, position. 
Moreover, these can be aligned with the AVDM response to be sent to hosting PC with a newly developed communication system. 
All of these components form the pre-requisite of this research.

\begin{figure}[t]
	\centering
	\subfloat[TF=0Hz, duration=30min]{\includegraphics[width=0.24\textwidth]{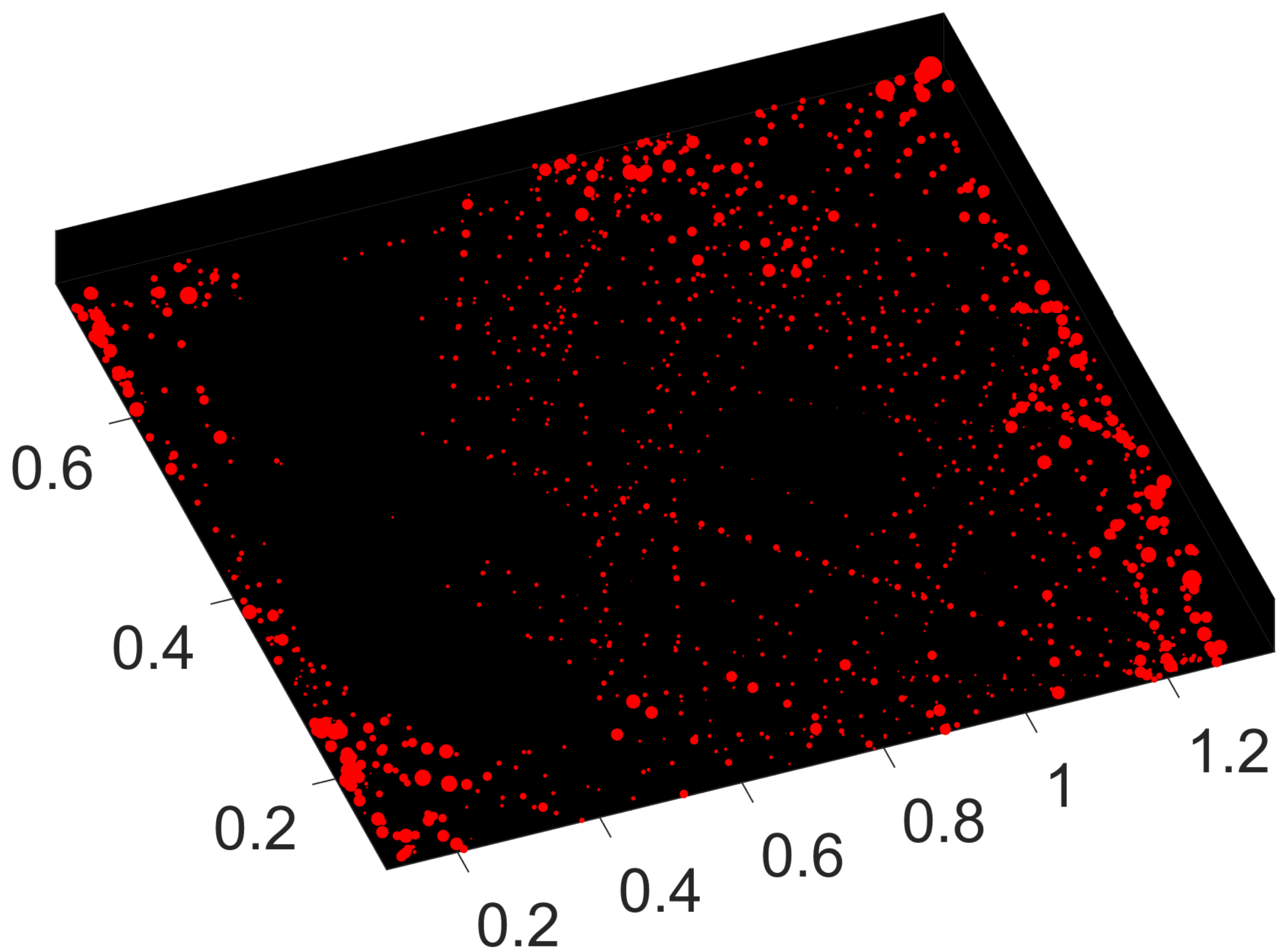}
		\label{Fig: profile-tf0-3D}}
	\hfil
	\subfloat[TF=0Hz, AVDM$>$100]{\includegraphics[width=0.22\textwidth]{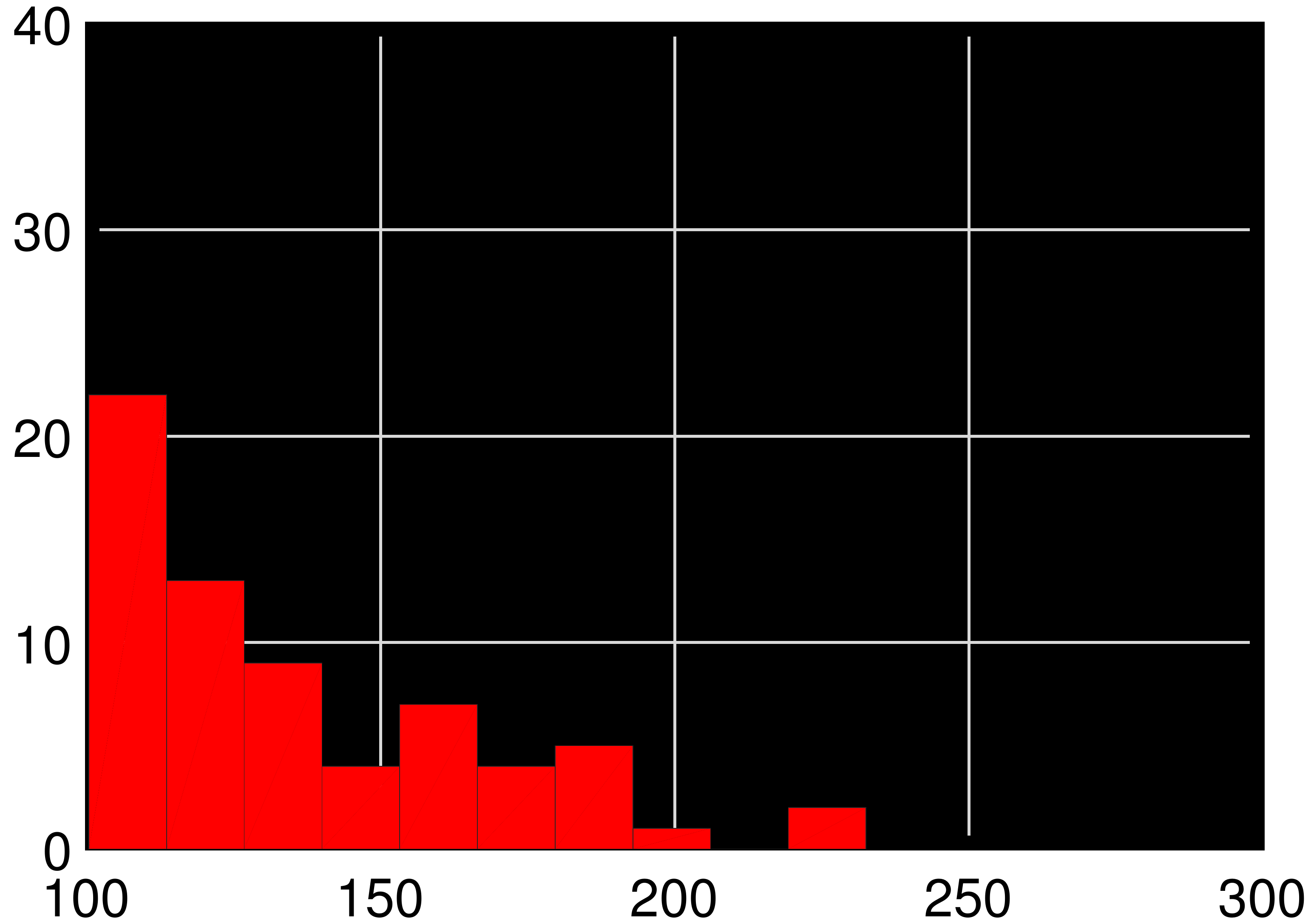}
		\label{Fig: hist-t0}}
	\vfil
	\vspace{-5pt}
	\subfloat[TF=1Hz, duration=30min]{\includegraphics[width=0.24\textwidth]{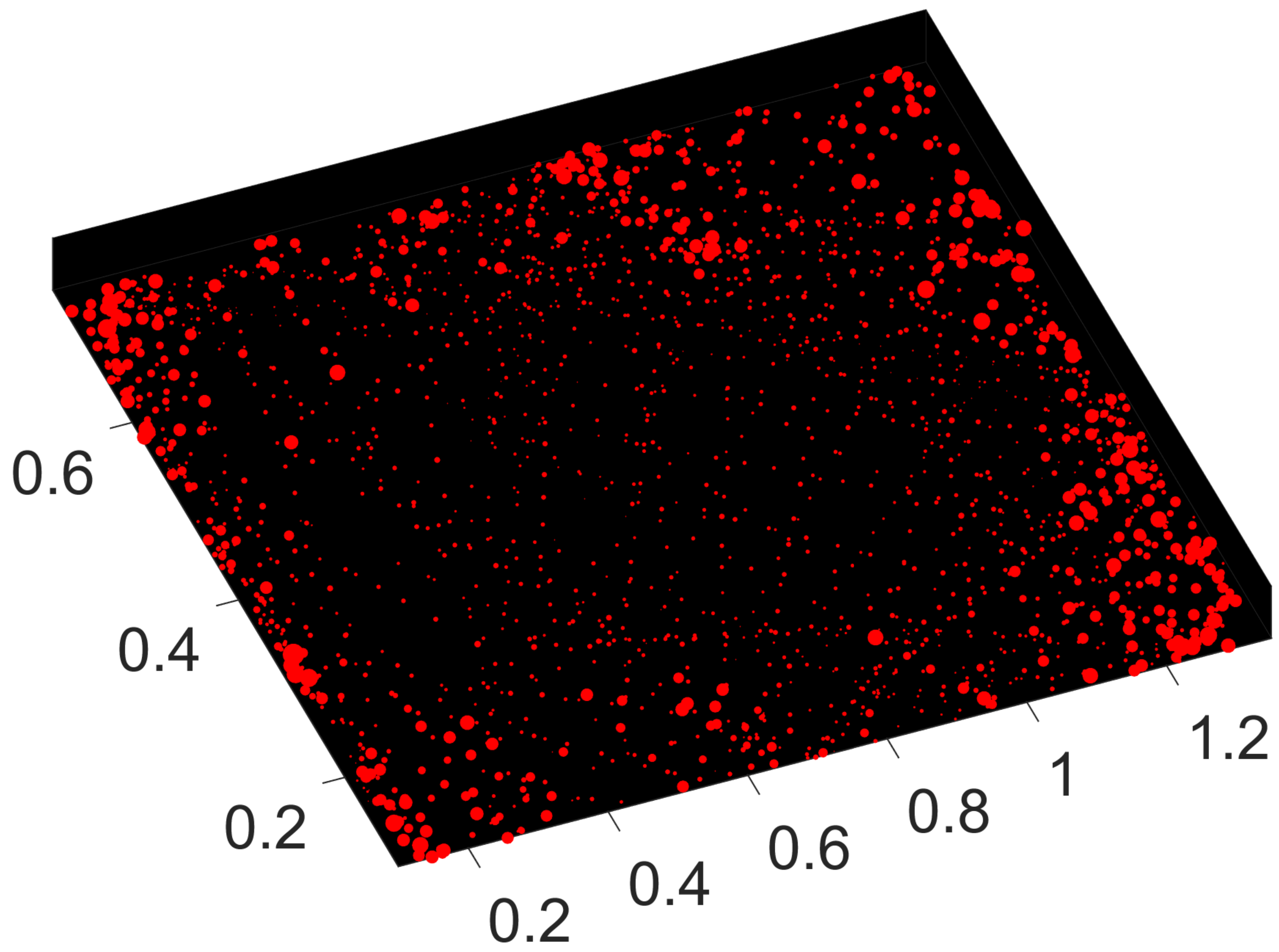}
		\label{Fig: profile-tf1-3D}}
	\hfil
	\subfloat[TF=1Hz, AVDM$>$100]{\includegraphics[width=0.22\textwidth]{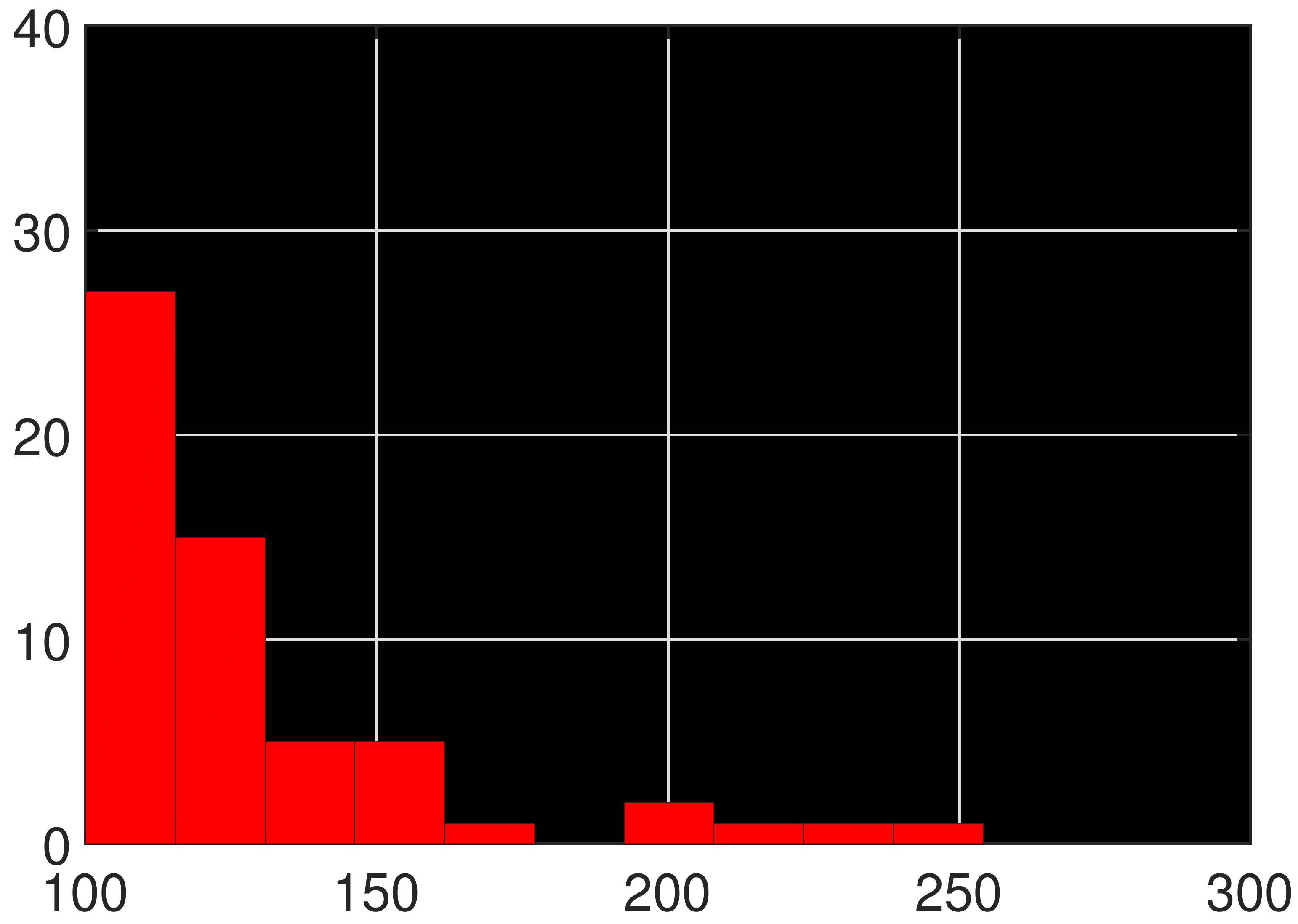}
		\label{Fig: hist-t1}}
	\vfil
	\vspace{-5pt}
	\subfloat[TF=3Hz, duration=30min]{\includegraphics[width=0.24\textwidth]{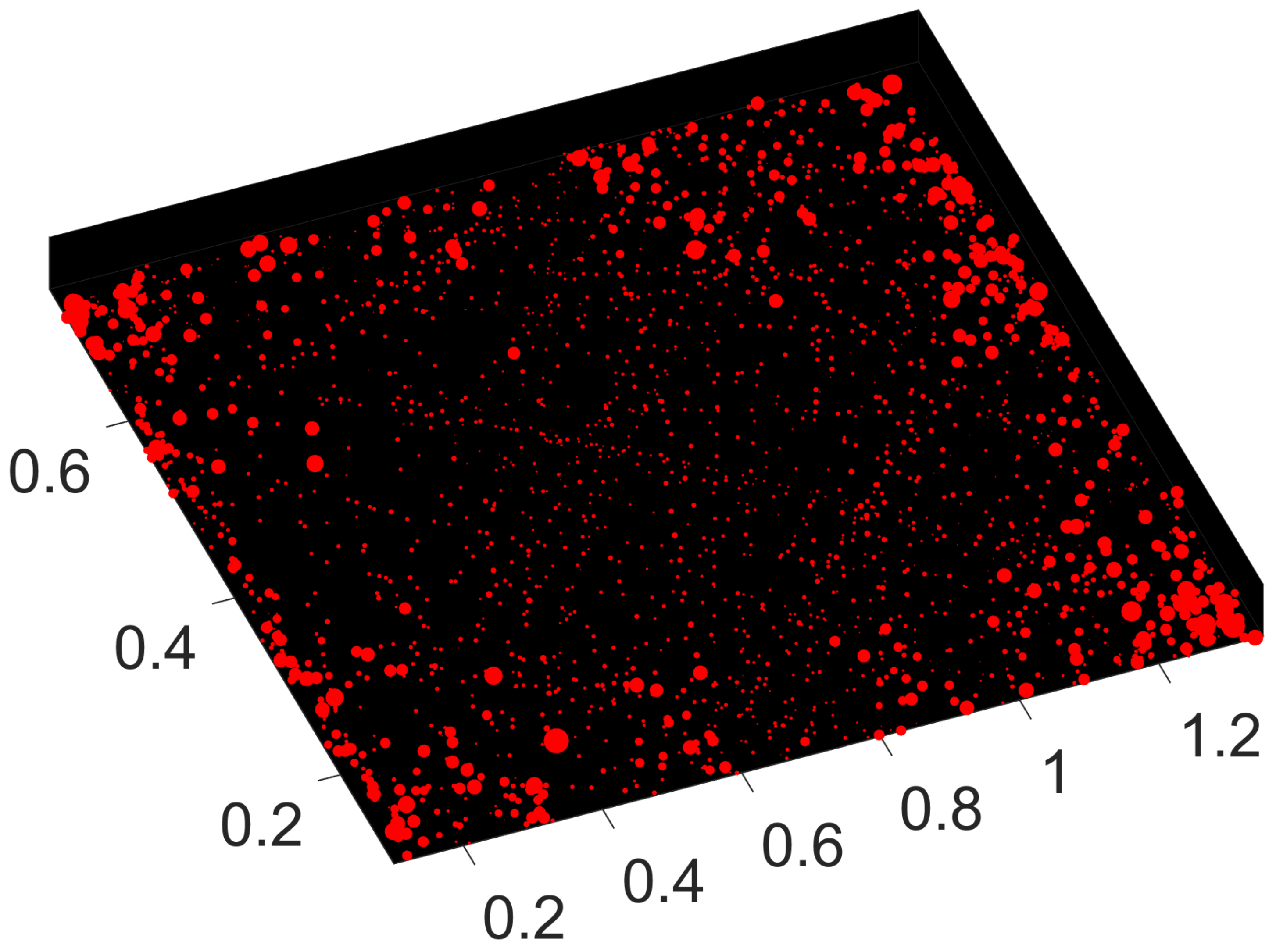}
		\label{Fig: profile-tf3-3D}}
	\hfil
	\subfloat[TF=3Hz, AVDM$>$100]{\includegraphics[width=0.22\textwidth]{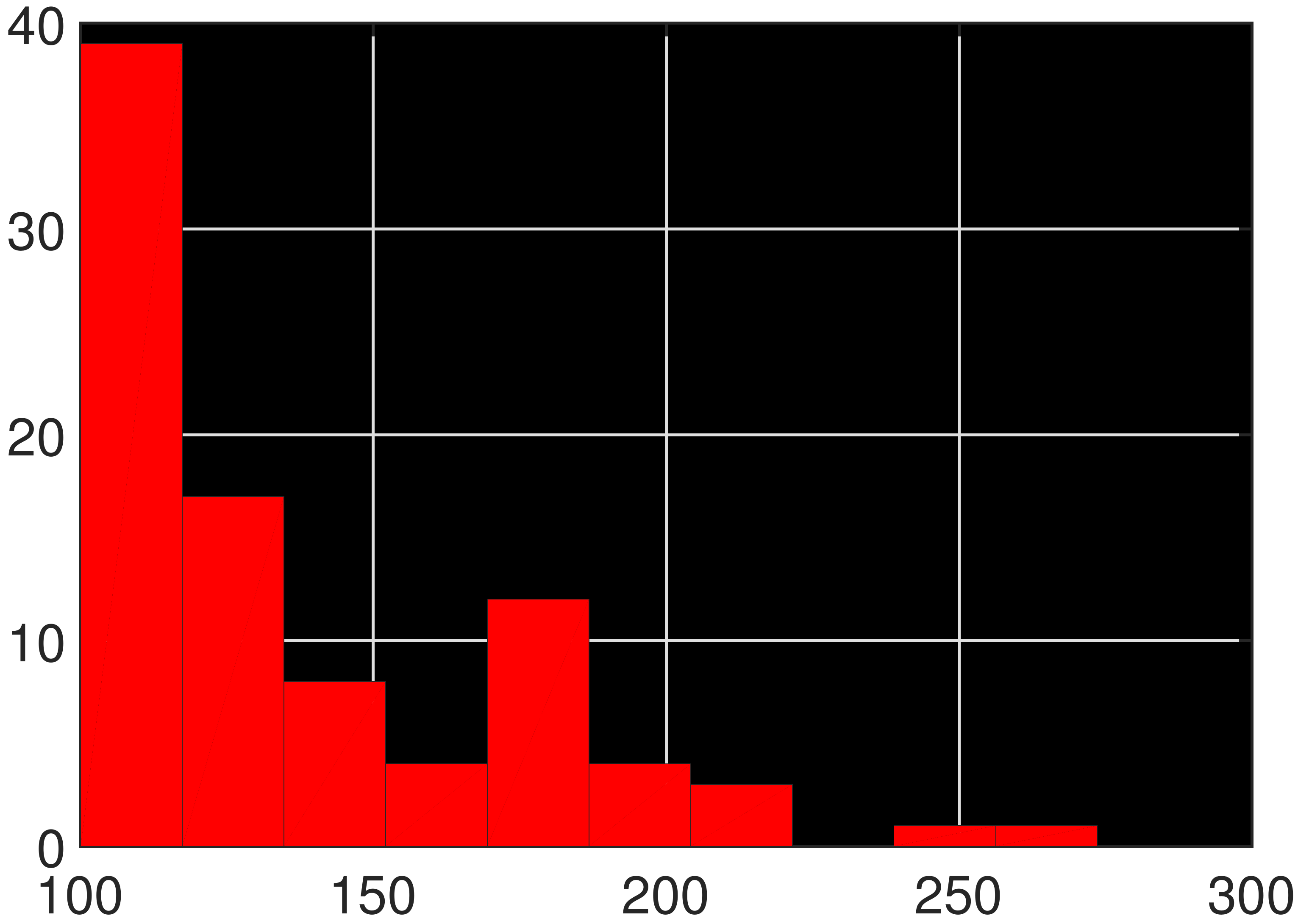}
		\label{Fig: hist-t3}}
	\caption{
		The profiled dynamic complexity of \textbf{grating visual scene} by robot overtime navigation. 
		The TF of grating scene was is at 0, 1, 3Hz, respectively. 
		The larger circles indicate the stronger model dynamic response with respect to time, i.e., the higher dynamic complexity of visual scene. 
		In the distribution histograms, the AVDM responses over 100 are counted.
	}
	\label{Fig: complexity profile grating}
	\vspace{-5pt}
\end{figure}

\begin{figure}[h!]
	\centering
	\subfloat[TF=0Hz, duration=30min]{\includegraphics[width=0.24\textwidth]{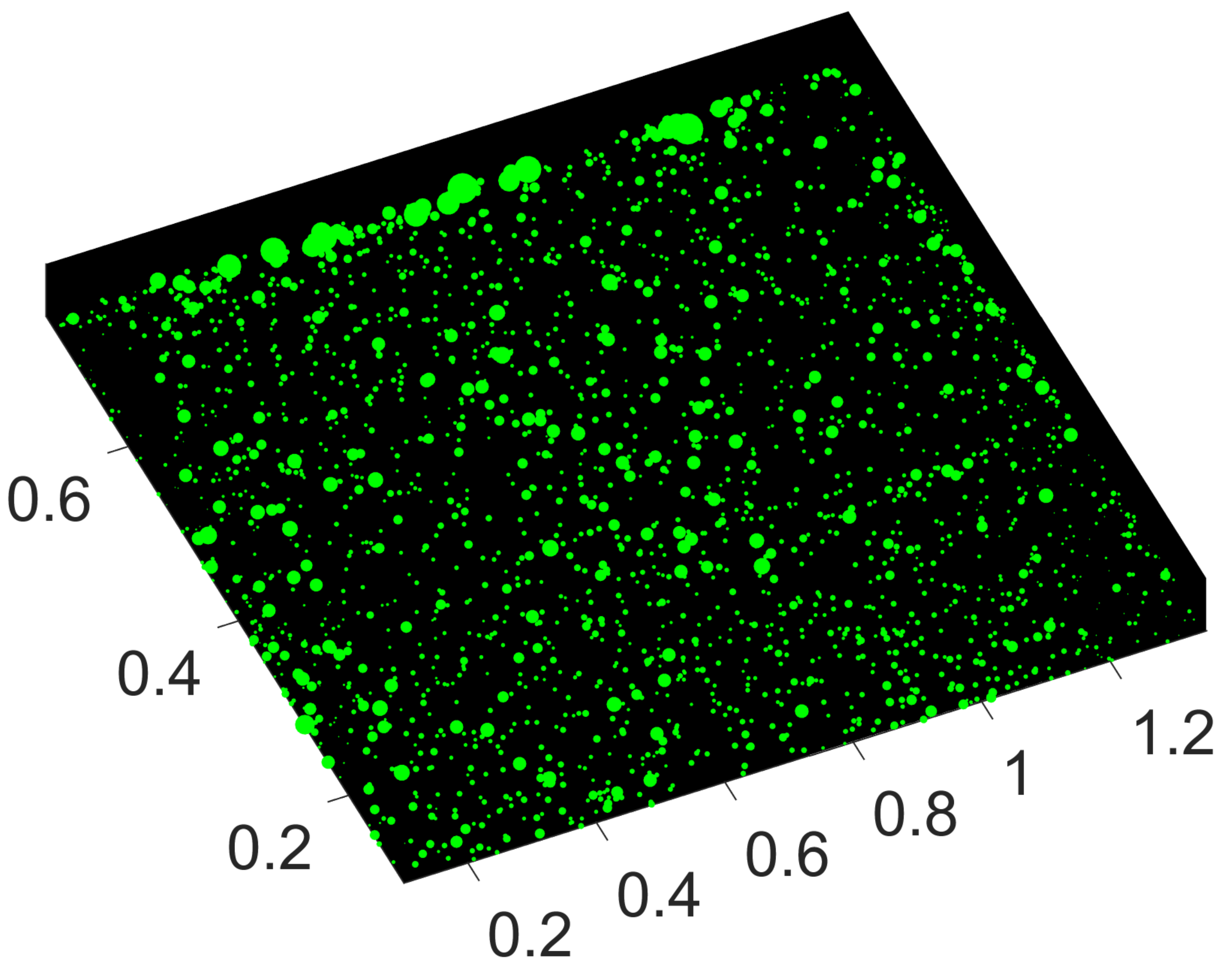}
		\label{Fig: profile-nature-tf0-3D}}
	\hfil
	\subfloat[TF=0Hz, AVDM$>$100]{\includegraphics[width=0.22\textwidth]{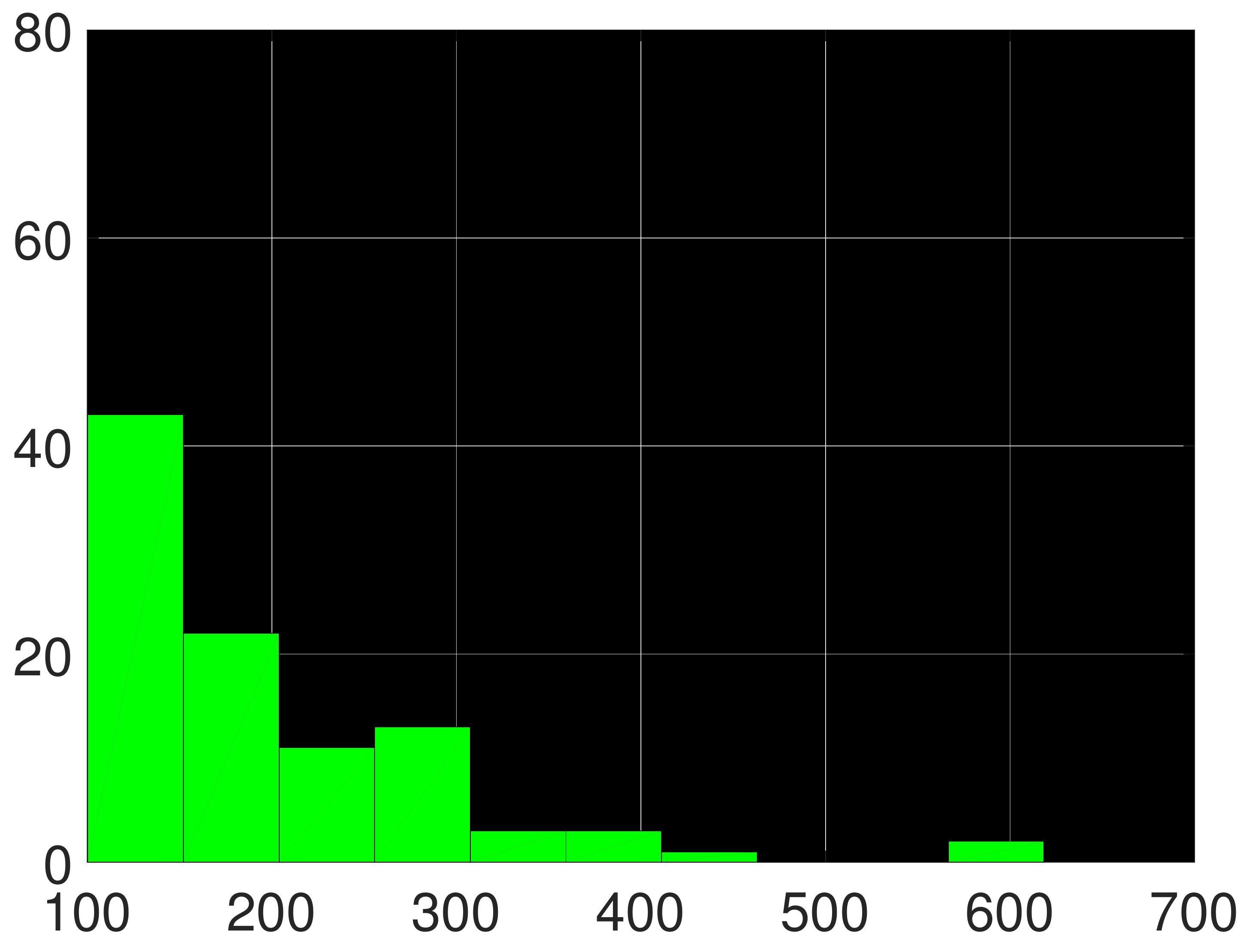}
		\label{Fig: hist-nature-t0}}
	\vfil
	\vspace{-5pt}
	\subfloat[TF=3Hz, duration=30min]{\includegraphics[width=0.24\textwidth]{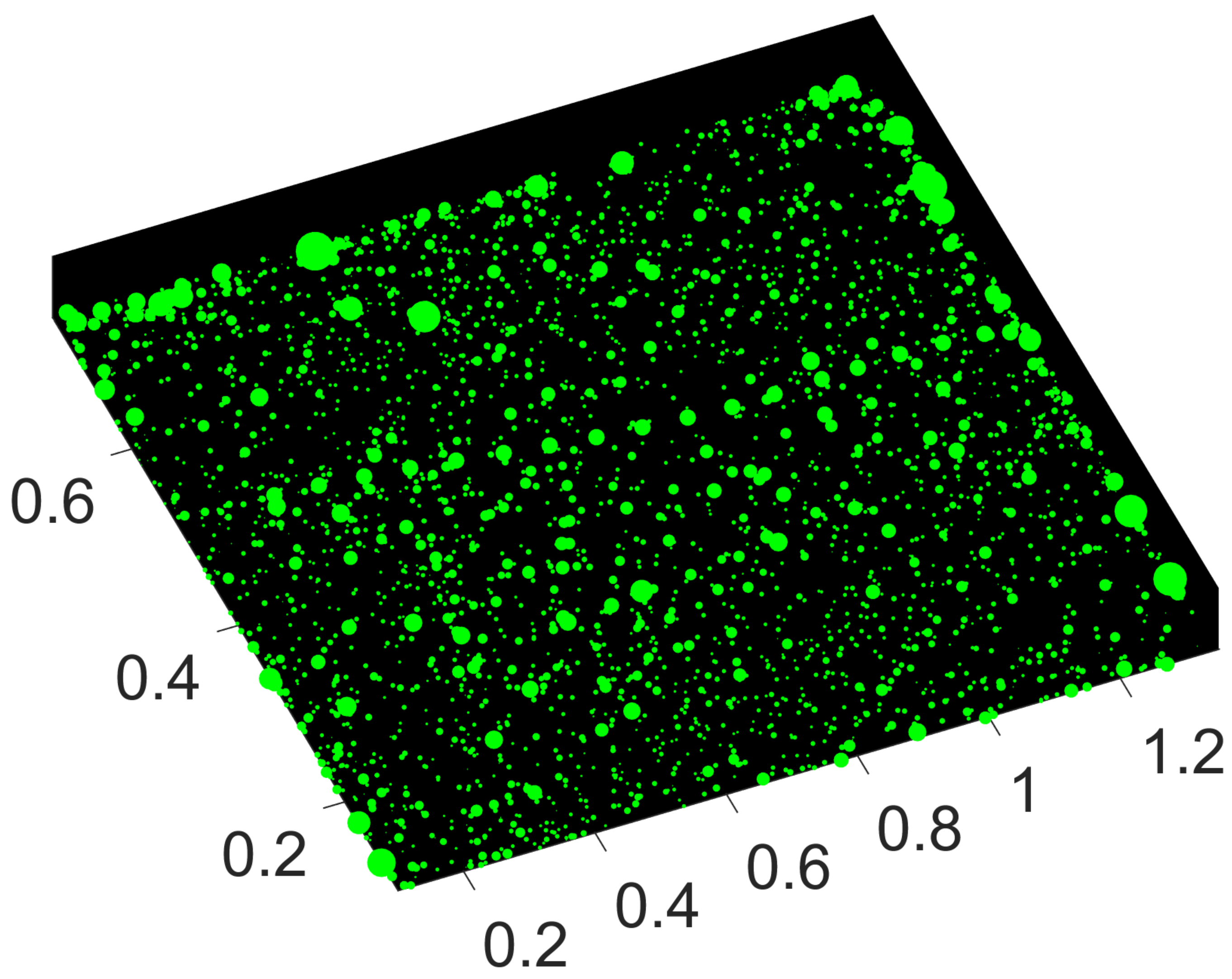}
		\label{Fig: profile-nature-tf3-3D}}
	\hfil
	\subfloat[TF=3Hz, AVDM$>$100]{\includegraphics[width=0.22\textwidth]{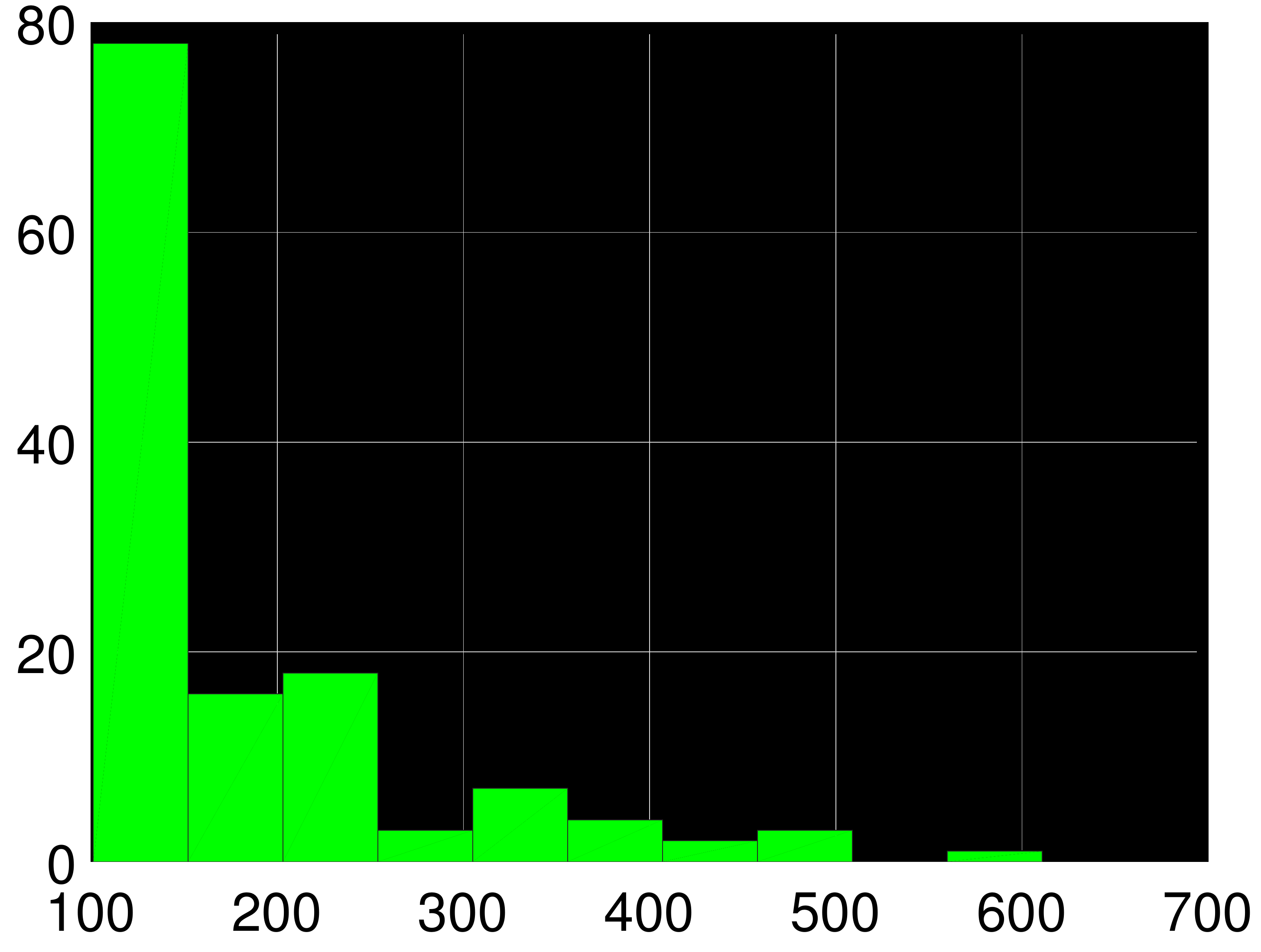}
		\label{Fig: hist-nature-t3}}
	\caption{
		The profiled dynamic complexity of \textbf{cluttered natural scene} by robot overtime navigation. 
		The TF of natural scene is set at 0 and 3Hz.
	}
	\label{Fig: complexity profile nature}
	\vspace{-10pt}
\end{figure}

\section{Metric Validation}
\label{Sec: evaluation}

The first step is to verify our hypothesis that such a biologically plausible visual processing scheme can reflect dynamic complexity in a quantitative manner. 
The complexity here is associated with SF-TF of visual scene. 
We carry out off-line simulations with input signal stimuli covering a wide range of SF-TF frequency sine-wave gratings (see examples in Fig. \ref{Fig: grating-stimuli}). 
Based on the Nyquist's law (sampling theorem), our sampling frequency of off-line video sequence is set at 300Hz; 
the investigated temporal frequency in this type of experiments is set below 150Hz. 
The resolution of videos is set at 100$\times$100 in pixels.

\begin{figure*}[t]
	\centering
	\subfloat[TF=0Hz, time=1$\sim$30min]{\includegraphics[width=0.32\textwidth]{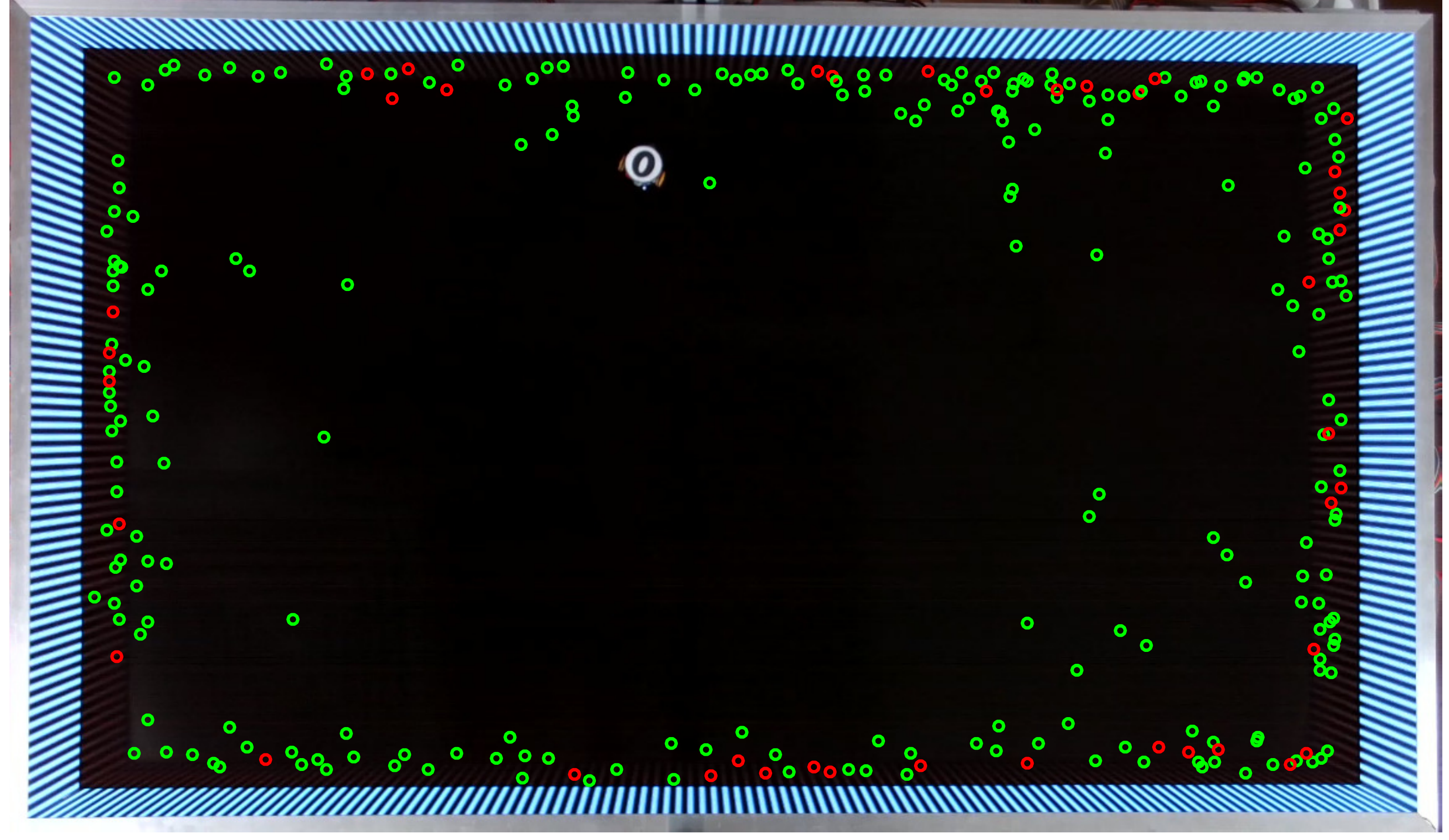}
		\label{Fig: cd-test-t0}}
	\hfil
	\subfloat[TF=3Hz, time=31$\sim$60min]{\includegraphics[width=0.32\textwidth]{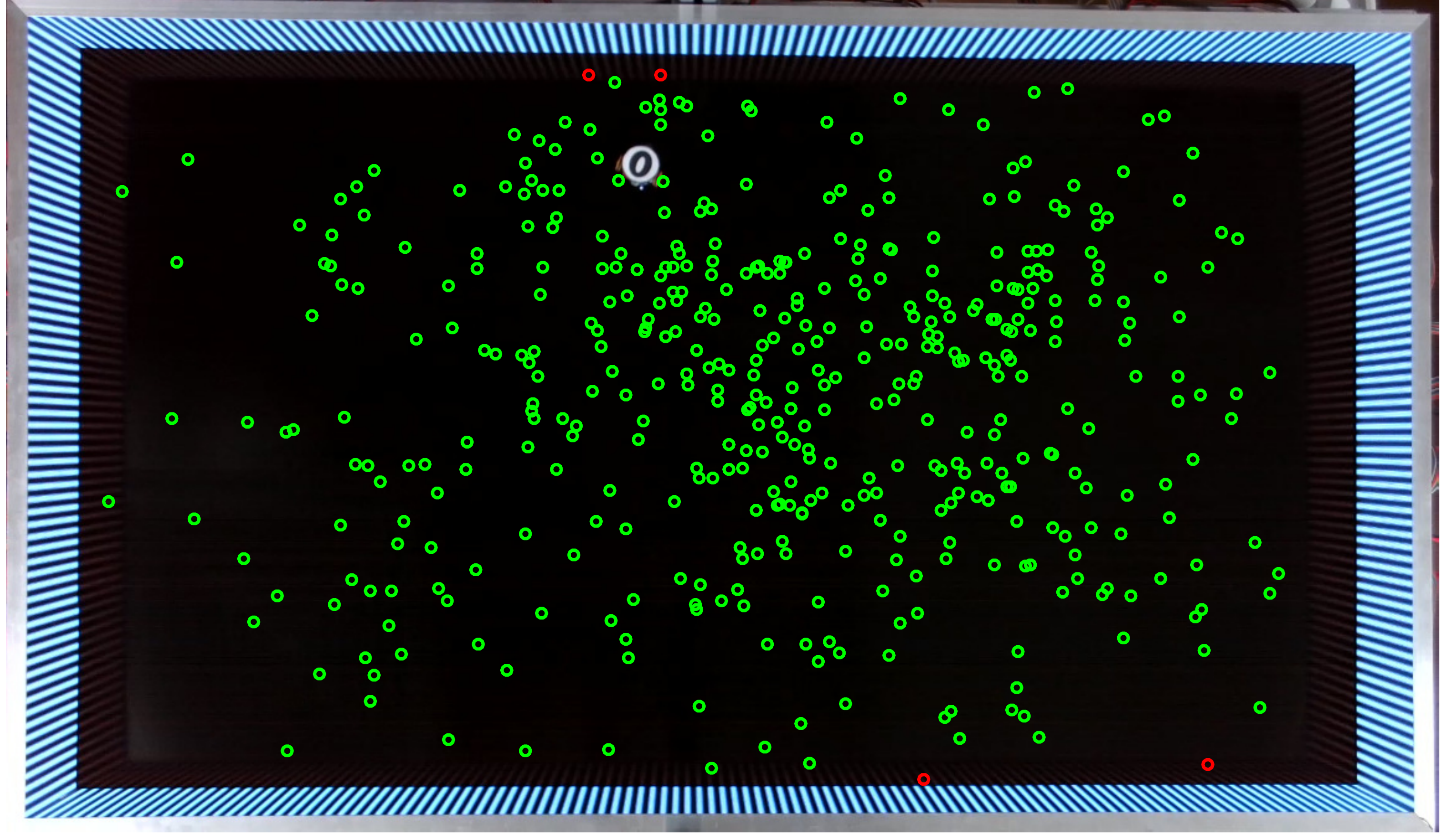}
		\label{Fig: cd-test-t3}}
	\hfil
	\subfloat[TF=3Hz, time=61$\sim$90min, \textbf{model adjusted}]{\includegraphics[width=0.32\textwidth]{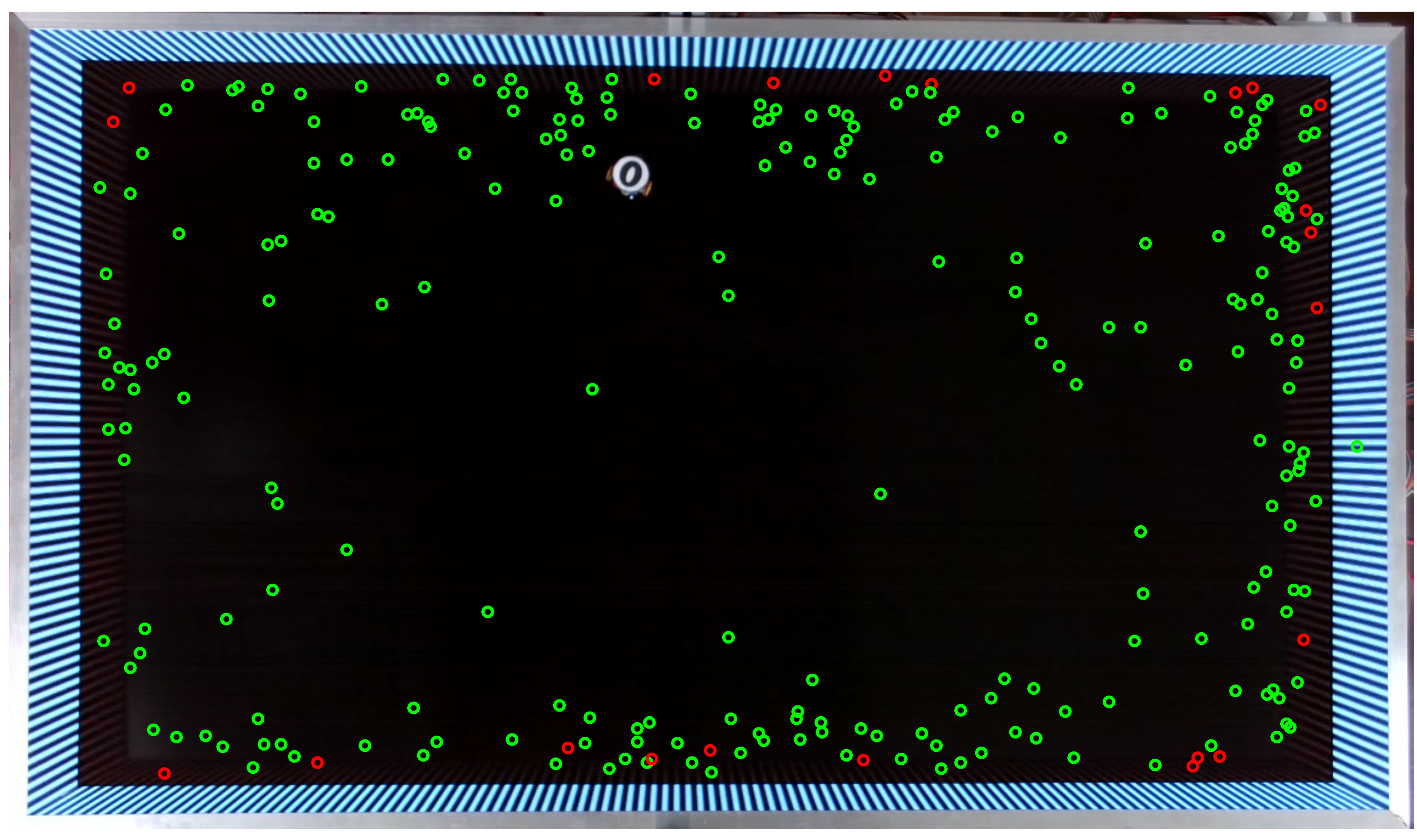}
		\label{Fig: cd-test-t3-2}}
	\vfil
	\vspace{-5pt}
	\subfloat[time=1$\sim$30, \textbf{229} avoidances, SR$>$90\%]{\includegraphics[width=0.325\textwidth]{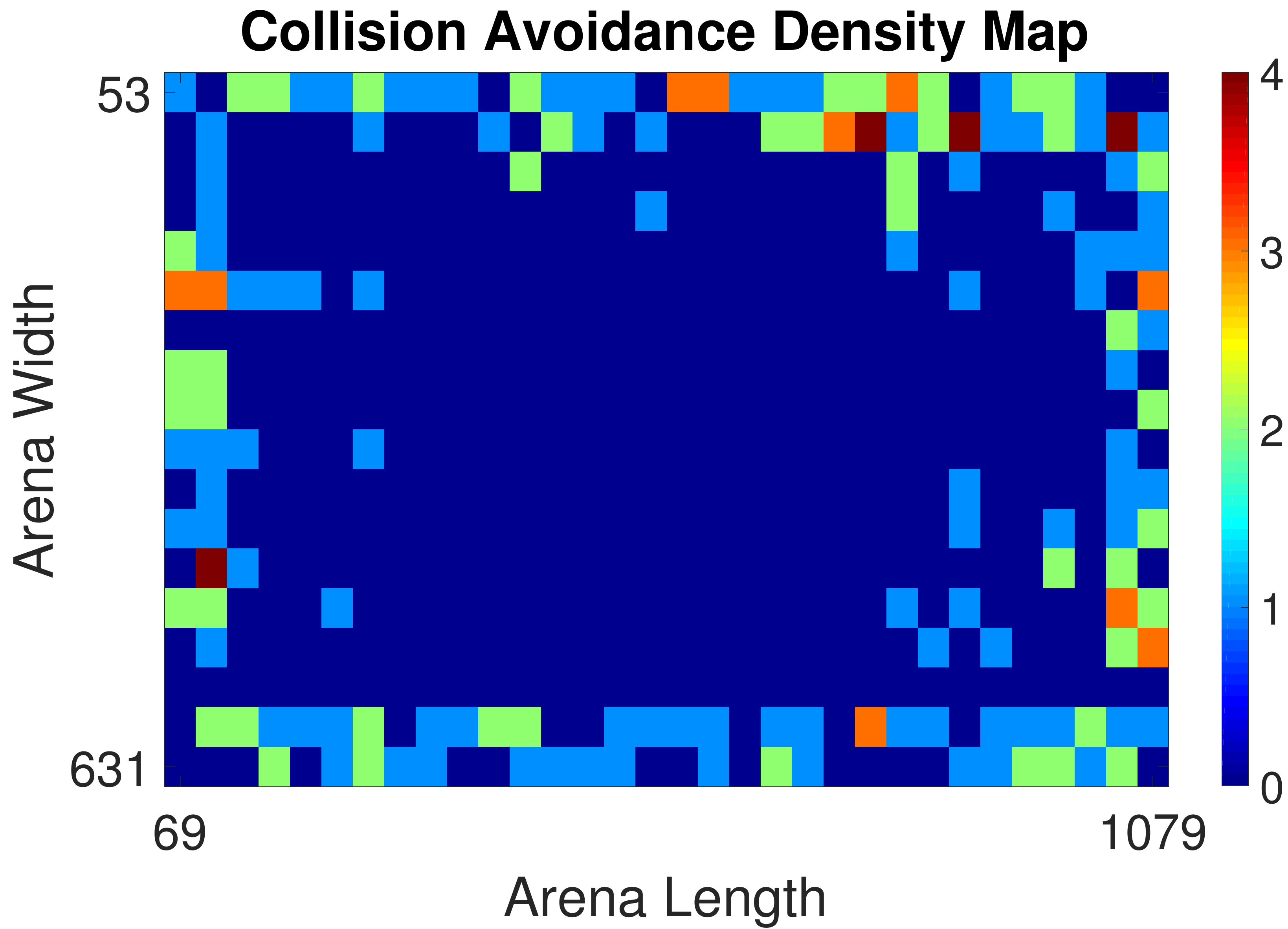}
		\label{Fig: cd-density-tf0}}
	\hfil
	\subfloat[time=31$\sim$60, \textbf{432} avoidances, SR$>$90\%]{\includegraphics[width=0.325\textwidth]{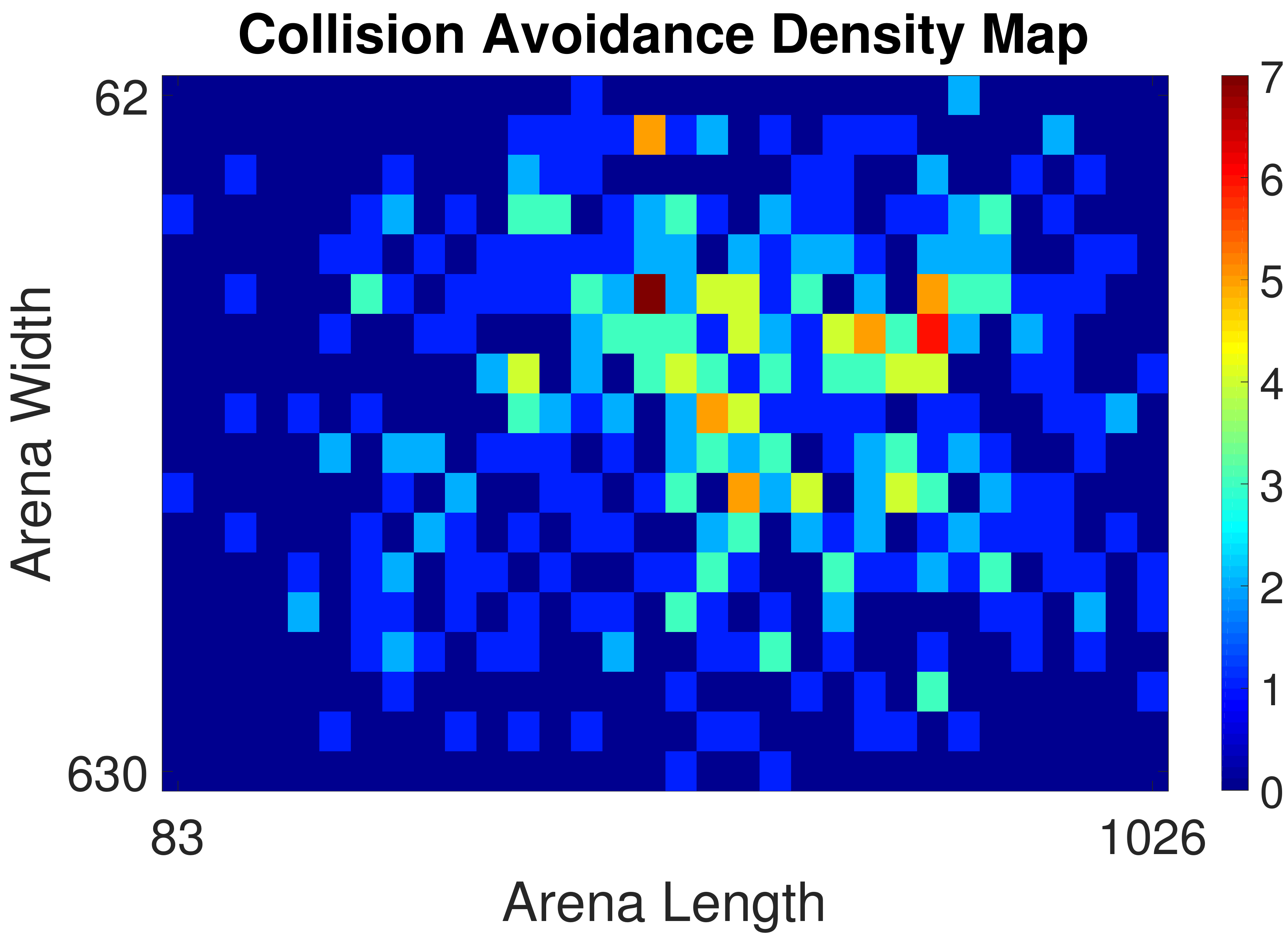}
		\label{Fig: cd-density-tf3}}
	\hfil
	\subfloat[time=61$\sim$90, \textbf{233} avoidances, SR$>$90\%]{\includegraphics[width=0.325\textwidth]{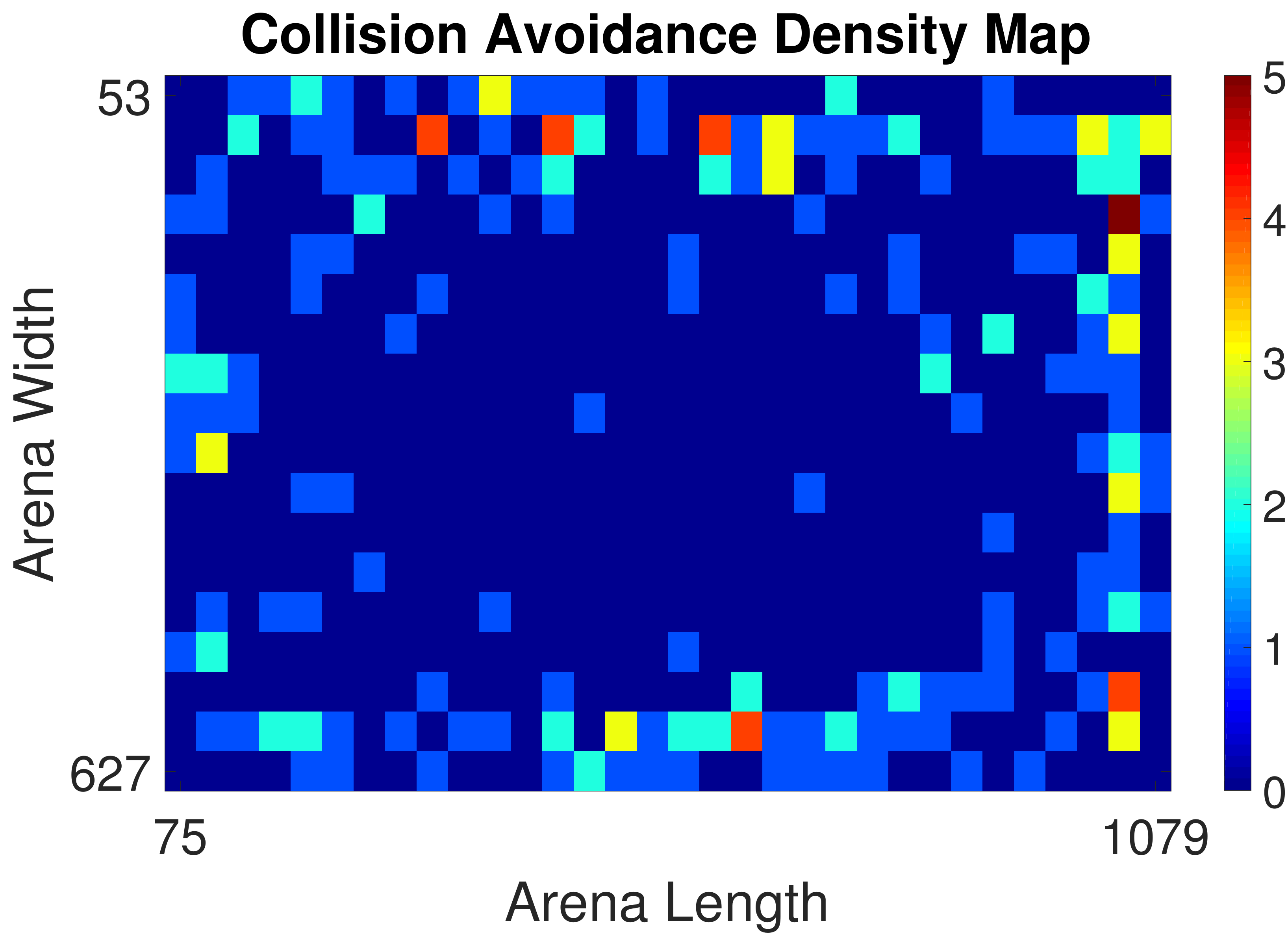}
		\label{Fig: cd-density-tf3-2}}
	\caption{
		Collision detection-and-avoidance density maps generated by robot visual navigation. 
		In (a),(b),(c), green circles indicate the avoidance positions whilst red circles denote the crashes on wall by failing detection. 
		Note that the proposed metric works as a predictor of surrounding visual dynamic complexity. 
		After perceiving such complexity, the tested model can be adjusted to leverage performance and energy consumption.
	}
	\label{Fig: collision detection task}
	\vspace{-10pt}
\end{figure*}

Fig. \ref{Fig: simulation results} compares the results of AVDM response by changing the ST or TF in a fairly wide range. 
The AVDM's dynamic response shows perfect monotonic increasing when the SF grows at each certain TF. 
When fixing the SF and increasing the TF, such a monotonic mapping generally aligns the former case though the dynamic response shrinks at some tested frequencies. 
As a result, a new metric herein can be established to describe the dynamic complexity in the context, that is, higher-frequency (either SF or TF) movements likely elicit stronger model dynamic responses. 
This can be used as a reasonable indicator or predictor for such an implicit attribute of visual scene.

To verify its scalability and efficacy, we design open-loop robot tests in real physical scene. 
To this end, the robot is initiated to move toward the visual wall displaying moving gratings with certain TF at 0, 1, 2, 3Hz, respectively, which resemble the circumstances in Fig. \ref{Fig: avdm-tf1}, \ref{Fig: avdm-tf5}, and \ref{Fig: avdm-tf50}. 
It can be clearly seen from Fig. \ref{Fig: robotic evaluation} the model dynamic response also generally represents the monotonic mapping to the contextual complexity despite fluctuations challenged by real physical grating wall. 
The robot evaluation demonstrates the proposed metric works consistently in real world visual scene.

\section{Profiling Dynamic Complexity}
\label{Sec: profile}

In this section, we profile the visual dynamic complexity. 
We carry on closed-loop experiments via robot visual navigation. 
Two types of on-line tests have been involved. 
Fig. \ref{Fig: arena-robot} and \ref{Fig: arena settings} show the arena settings including grating and cluttered natural visual scenes. 
The robot motion is bound within the field of white-light ring displayed by the TV at bottom, near the surround visual walls.  
Technically speaking, this strategy is used for invoking the aforementioned ``turning response". 
The light ring can be picked up by the robot's light sensors assembled at bottom to turn around, in order to avoid any human interventions in visual navigation.

In the first type of grating scene tests, Fig. \ref{Fig: complexity profile grating} exhibits the profiled complexity distributed in the arena, against different TF of grating wall. 
Intuitively, Fig. \ref{Fig: profile-tf0-3D}, \ref{Fig: profile-tf1-3D}, and \ref{Fig: profile-tf3-3D} all articulate that the model dynamic response is more intensely aroused near the edges of arena indicating higher dynamic complexity in the context. 
The results are consistent with the previous simulation and robot open-loop tests. 
Furthermore, Fig. \ref{Fig: hist-t0}, \ref{Fig: hist-t1}, and \ref{Fig: hist-t3} compares the magnitude of model dynamic responses. 
Interestingly, the profiled complexity becomes fairly higher by increasing the TF of drifting grating scene, as the neural activity is stronger as illustrated in the corresponding distribution.

In the second type of natural scene tests, the complexity of more realistic visual scene is explicitly profiled. 
Fig. \ref{Fig: complexity profile nature} compares the nature scene complexity against two TF of drifting visual scene. 
The results can be aligned with the former grating scene tests which demonstrate the proposed method could be scalable to more challenging, realistic environment. 
Accordingly, the visual dynamic complexity has been quantitatively estimated and explicitly described.

\section{Using This Metric}
\label{Sec: investigation}

In this section, we demonstrate a case study on using this metric. 
As introduced in Section \ref{Sec: introduction}, dynamic complexity is a hidden attribute of visual scene that is very difficult to be described explicitly, and could affect accurate assessment of dynamic vision systems especially in real world scenes. 
The current literature lacks effective methods from both spatial and temporal perspectives. 
Therefore, the visual system models are likely overestimated with limited testing data described ambiguously as ``complex dynamic visual scene".

Through this research, we have put forward a new metric which can alleviate this situation. 
For example, the boundary of visual system models can be plainly found by quantitatively increasing such complexity. 
Using this metric, here we investigate a collision detection visual system recently proposed by Fu et al. \cite{Fu-2020-Access}. 
The robot in this study is applied to implement this model interacting with surrounding visual walls for collision detection and avoidance in navigation. 
Note that the motion control is not the focus of this paper. 
Accordingly, the robot avoidance behaviour is set simply to turn randomly to right or left at 80 $\sim$ 100 degrees. 
With respect to previous research experience, two classic metrics on assessing collision detection visual systems have been considered: 
(1) The success rate (SR) of collision detection-and-avoidance indicates the system robustness which should be maintained at an acceptable level ($>$90\% with regard to recent studies using the micro-robot \cite{Fu-2020-Access}). 
(2) The distance to collision (DTC) means how far the potential risk is detected and avoided which indicates the system sensitivity; an appropriate DTC can reduce energy consumption in confined space that is indicated by the amount of avoidances in an identical time window in this case study.

This experiment involves three procedures. 
Fig. \ref{Fig: collision detection task} illustrates the robot crash and avoidance events with density maps in the arena. 
Initially, the TF of grating wall is set at 0Hz, most of the successful avoidances happened near the peripheries of arena (Fig. \ref{Fig: cd-test-t0} and \ref{Fig: cd-density-tf0}). 
The DTC is relatively low and the SR is maintained above the satisfactory level with 229 avoidances in total. 
Under the same set of model parameters, we increase the complexity of visual scene by setting the TF at 3Hz. 
Interestingly, Fig. \ref{Fig: cd-test-t3} and \ref{Fig: cd-density-tf3} show obviously that the avoidance event density is much more centralised toward the centre field of arena, in contrast with the performance in relatively simpler visual scene. 
The tested model becomes more sensitive. 
In this case, the DTC significantly extends with more energy consumption. 
Accordingly, a boundary of this tested model has been found in visual scene with increased dynamic complexity.

Such a metric works effectively as a predictor of surrounding input complexity to visual systems which can explain, and be coupled with the different behaviours. 
After perceiving such complexity, a next interesting attempt can be finding an either learning or experience based adjustment to make the tested visual system adapt to the new environment with performance and power consumption leveraged. 
Fig. \ref{Fig: cd-test-t3-2} and \ref{Fig: cd-density-tf3-2} together show a satisfactory balance by simply increasing the threshold for collision detection. 
To sum up, this case study reveals a promising way of using the proposed metric to inform the agent surrounding dynamic complexity to make sensible adjustment or choice.

\section{Conclusion}
\label{Sec: conclusion}

This paper has presented a novel bio-robotic approach to estimate and profile visual dynamic complexity that is essentially a concealed, intractable attribute of visual scene. 
This approach has integrated biologically plausible visual processing scheme and robotic techniques to set up a metric describing such complexity explicitly and quantitatively, from both spatial and temporal perspectives. 
We have progressed this research with several steps. 
Firstly, we have shown a method to fill the gap of measuring complexity associated with spatial and temporal frequency attributes of visual scene. 
To establish this metric, we have then validated it through off-line simulation and on-line bio-robot tests. 
Importantly, we have found a monotonic mapping from the neural dynamic response to a wide range of frequencies of drifting visual scene. 
Next, we have applied an autonomous micro-mobile robot to profile the dynamic complexity in visual navigation to describe surrounding input complexity within an arena with visual walls displaying drifting visual scene. 
This approach has also been verified to be flexible and effective in more realistic, natural visual scene. 
Finally, we have given a case study of using this metric as a predictor of surrounding visual dynamic complexity to help find the boundary of a collision detection visual system in more complex dynamic environment.

Our future works are 1) consolidating this metric with improvement in dynamic visual processing formula to cope with various realistic natural scenes; 2) coupling this predictor of input complexity with learning methods to benefit other dynamic vision systems; 3) introducing this work into relevant robot visual tasks to help the robot measure, and adapt to changing environment.

\section*{Acknowledgement}

This research has received funding from the National Natural Science Foundation of China under the Grant No 12031003, the China Postdoctoral Science Foundation Grant 2019M662837, 2020M682651, and the European Union's Horizon 2020 research and innovation programme under the Marie Sklodowska-Curie Grant Agreement No 778602 ULTRACEPT.

\section*{Appendix}

We have a supplementary video demo with this paper for the description of methods.

\bibliographystyle{IEEEtran}
\bibliography{qinbingbib}

\end{document}